\PassOptionsToPackage{table}{xcolor}
\documentclass[sigconf,screen]{acmart}

\AtBeginDocument{%
  }

\copyrightyear{2026}
\acmYear{2026}
\setcopyright{cc}
\setcctype{by}
\acmConference[MM '26] {Proceedings of the 34th ACM International Conference on Multimedia}{November 10--14, 2026}{Rio de Janeiro, Brazil.}
\acmBooktitle{Proceedings of the 34th ACM International Conference on Multimedia (MM '26), November 10--14, 2026, Rio de Janeiro, Brazil}
\acmISBN{979-8-4007-2213-4/2026/11}
\acmDOI{10.1145/3767308.3835174}
\usepackage{pifont}
\usepackage{multirow}
\usepackage{booktabs}
\usepackage{tabularx}
\usepackage{placeins}
\usepackage{enumitem}
\newcommand\rankfirst[1]{\textbf{#1}}
\newcommand\ranksecond[1]{\underline{#1}}
\newcommand{\mainresultgroup}[1]{\rowcolor{black!8}\multicolumn{17}{c}{\textit{#1}} \\}
\newcommand{\cmark}{\ding{51}}
\newcommand{\xmark}{\ding{55}}

\definecolor{lightgreen}{RGB}{220,245,220}
\definecolor{lightred}{RGB}{255,235,235}

\begin{document}

\title{One Patch Is Enough: Reinforcement-Optimized Visual Token Grounding for MLLM-Based Scene Text Spotting}

\author{Rui Tang}
\email{eeruitang@mail.scut.edu.cn}
\affiliation{%
  \institution{South China University of Technology}
  \city{Guangzhou}
  \country{China}}

\author{Wentao Yang}
\email{wente\_young@foxmail.com}
\affiliation{%
  \institution{South China University of Technology}
  \city{Guangzhou}
  \country{China}}
  \affiliation{%
  \institution{HiThink Research}
  \city{Hangzhou}
  \country{China}}

\author{Peirong Zhang}
\email{eeprzhang@mail.scut.edu.cn}
\affiliation{%
  \institution{South China University of Technology}
  \city{Guangzhou}
  \country{China}}

\author{Yongxin Shi}
\email{yongxin\_shi@foxmail.com}
\affiliation{%
  \institution{South China University of Technology}
  \city{Guangzhou}
  \country{China}}

\author{Shun Zhang}
\email{scutzhangs@163.com}
\affiliation{%
  \institution{South China University of Technology}
  \city{Guangzhou}
  \country{China}}

\author{Huiguo He}
\email{hehuiguo@scut.edu.cn}
\affiliation{%
  \institution{South China University of Technology}
  \city{Guangzhou}
  \country{China}}

\author{Lianwen Jin}
\authornote{Corresponding author.}
\email{eelwjin@scut.edu.cn}
\affiliation{%
  \institution{South China University of Technology}
  \city{Guangzhou}
  \country{China}}

\renewcommand{\shortauthors}{Rui Tang et al.}

\begin{abstract}
Scene text spotting requires high-precision alignment between textual recognition and spatial localization. While visual-token grounding has emerged as a promising formulation for Multimodal Large Language Models (MLLMs), the previous multi-patch paradigm often introduces redundant noise and localization ambiguity, particularly for dense or small text instances. To address this, we propose \textbf{Single-Patch Text Spotting (SPaTS)}, a vision-centric framework that routes each text instance through a \emph{single} anchor visual token and then recovers geometry via full-image refinement. 
To accurately identify this anchor without oracle labels, we introduce \textbf{Single-Patch Selective Optimization (SPaSO)}, a reinforcement learning framework that optimizes discrete visual-token selection using patch-level rewards. To further improve representation robustness and localization precision, we introduce \textbf{Directional Embedding Alignment (DEA)} to suppress unstable norm bias by decoupling feature magnitude and direction, and \textbf{Patch-Enhanced Decoding (PED)} to fuse the routed anchor with language semantics and cross-attend over the full-image feature map for geometry-aware boundary regression beyond coordinate-space surrogates. Extensive experiments demonstrate that SPaTS consistently and significantly outperforms both frontier closed-source MLLMs and OCR MLLMs. Code is available at \url{https://github.com/eeNickTang/SPaTS}.
\end{abstract}



\maketitle

\begin{figure}[h]
  \centering
  \includegraphics[width=\linewidth]{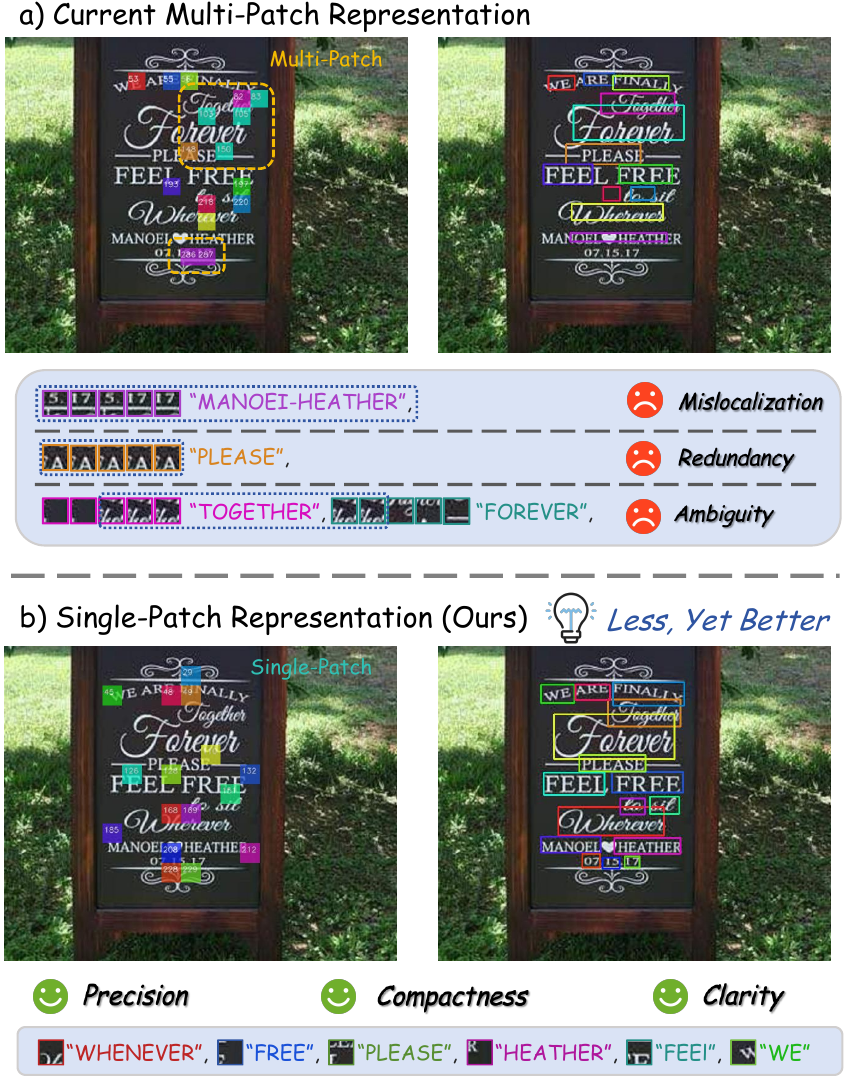}
  \caption{Motivation for SPaTS. In current multi-patch grounding, \emph{mislocalization} refers to irrelevant background patches, \emph{redundancy} refers to multiple patches assigned to the same text instance, and \emph{ambiguity} refers to mixed evidence from neighboring words or regions.}
  \vspace{-20pt}
  \label{fig:motivation}
  \Description{Motivation figure contrasting multi-patch grounding with the single-patch grounding strategy used in SPaTS.}
\end{figure}

\vspace{-5pt}

\section{Introduction}

Scene text spotting requires simultaneous text recognition and spatial grounding of each transcription to its corresponding image region. Specialized spotters have evolved from end-to-end pipelines that jointly optimize detection and recognition \cite{liu2018fots, liao2020real} to arbitrary-shape frameworks based on segmentation, curve regression, and query decoding \cite{lyu2018mask, liu2020abcnet, zhang2022testr, ye2023deepsolo, huang2022swintextspotter,liu2023sptsv2}. These expert models achieve strong geometric precision, yet their reasoning capability remains task-specific and constrained to predefined spotting workflows. Multimodal Large Language Models (MLLMs), conversely, provide semantic understanding and flexible generation \cite{jia2025visual, zhu2025internvl3, bai2025qwen3vl}, but lack the fine-grained spatial control requisite for precise localization.

Visual Token Grounding (VTG) provides a bridge between language generation and spatial localization. Methods such as Shikra~\cite{chen2023shikra} and Kosmos-2~\cite{peng2023kosmos} encode grounding through coordinate tokens~\cite{lggpt2025zhang}. Later approaches, including Groma~\cite{ma2024groma} and Osprey~\cite{yuan2024osprey}, replace coordinates with region proxies or mask-like abstractions. More recent methods, such as ClawMachine~\cite{ma2025clawmachine} and PaDT~\cite{su2026patch}, move one step closer to native visual evidence by grounding on image-derived patch tokens. This evolution is encouraging, but it also leaves a key question: \emph{what granularity of visual evidence is most appropriate for each grounding decision?}

The appropriate visual granularity is inherently domain dependent~\cite{Fang2024PUMA}. For generic objects, region-level or relatively coarse visual evidence is often sufficient to establish instance identity~\cite{ren2017faster}. In contrast, scene text instances are typically small in scale and embedded in complex scenes, where neighboring patches are more susceptible to spatial ambiguity due to adjacent words and background clutter~\cite{long2021scene, yang2019scrdet}. As illustrated in Fig.~\ref{fig:motivation}(a), aggregating multiple patches may entangle informative cues with irrelevant or misleading signals, leading to coupled errors in recognition and localization and introducing significant ambiguity in scene text spotting. Therefore, a natural question is whether a single informative patch is sufficient for further grounding, given that aggregating additional patches may introduce noise and exacerbate ambiguity~\cite{Ma2026One2SeqOW,zhang2025llavaonevision}.

To address these limitations, we propose \textbf{Single-Patch Text Spotting (SPaTS)}, a paradigm that routes each text instance through a \emph{single} anchor visual token. Instead of aggregating multiple patches at the language-model interface, SPaTS selects one anchor patch per instance and then delegates geometry recovery to PED, which cross-attends over the full-image feature map. In other words, SPaTS is a \emph{single-anchor grounding with full-image geometric refinement} design rather than a single-patch-only decoder. As shown in Fig.~\ref{fig:motivation}(b), this formulation yields a more compact routing signal and preserves a tighter coupling between the predicted geometry and the target image region, thereby reducing recognition-localization drift.

This definition presents a critical optimization challenge: patch informativeness is a \emph{latent attribute} lacking explicit ranking, making it difficult for supervised learning (SL) to identify the most discriminative evidence~\cite{lei2016rationalizing}. 
Since the optimal patch is only revealed by final spotting quality, this problem is naturally suited to reinforcement learning, which directly optimizes sampled discrete decisions using delayed task-level rewards~\cite{ouyang2022training,shao2024deepseekmath,liu2025visual_rft,shen2025vlm_r1}. 
We therefore reformulate patch selection as a discrete decision problem with task-level feedback, and introduce \textbf{Single-Patch Selective Optimization (SPaSO)}, a reinforcement learning framework that directly optimizes patch selection using patch-level rewards derived from the final spotting outcome while jointly optimizing the continuous decoder. In this way, SPaSO learns \emph{which} patch to select, while continuous supervision learns \emph{how} to regress its geometry.

To endow the single patch with robust representations across heterogeneous grounding formats and facilitate tighter alignment with linguistic semantics, we further address the architectural integration of the selected patch features. First, to stabilize selected patch features in the unified head, we introduce \textbf{Directional Embedding Alignment (DEA)}, which decouples feature magnitude and direction. This design suppresses unstable norm bias in patch competition and makes the selected representation focus more on directional alignment with the language state. Second, we propose \textbf{Patch-Enhanced Decoding (PED)} to alleviate the information bottleneck of decoding from a selected patch alone by fusing the selected visual patch with language hidden states, so the decoder can exploit richer visual-semantic cues for geometry prediction.

In summary, the main contributions of this work are as follows:
\begin{itemize}[leftmargin=*]
    \item We introduce \textbf{SPaTS}, a single-anchor visual-token paradigm for scene text spotting. It replaces noisy multi-patch routing with one anchor token per instance and full-image geometric refinement, leading to more compact and less ambiguous grounding.
    \item We propose \textbf{SPaSO}, a patch-level reinforcement learning framework that directly optimizes discrete visual-token selection when oracle optimal labels are unavailable, yielding state-of-the-art results across multiple benchmarks.
    \item We propose \textbf{DEA} and \textbf{PED}. DEA decouples patch magnitude and direction, using warmup to calibrate magnitude and later training to focus on directional alignment, while PED combines routed local evidence with language semantics to recover precise boundaries without reintroducing multi-patch noise.
\end{itemize}

\begin{figure*}[t!]
  \centering
  \includegraphics[width=0.97\textwidth]{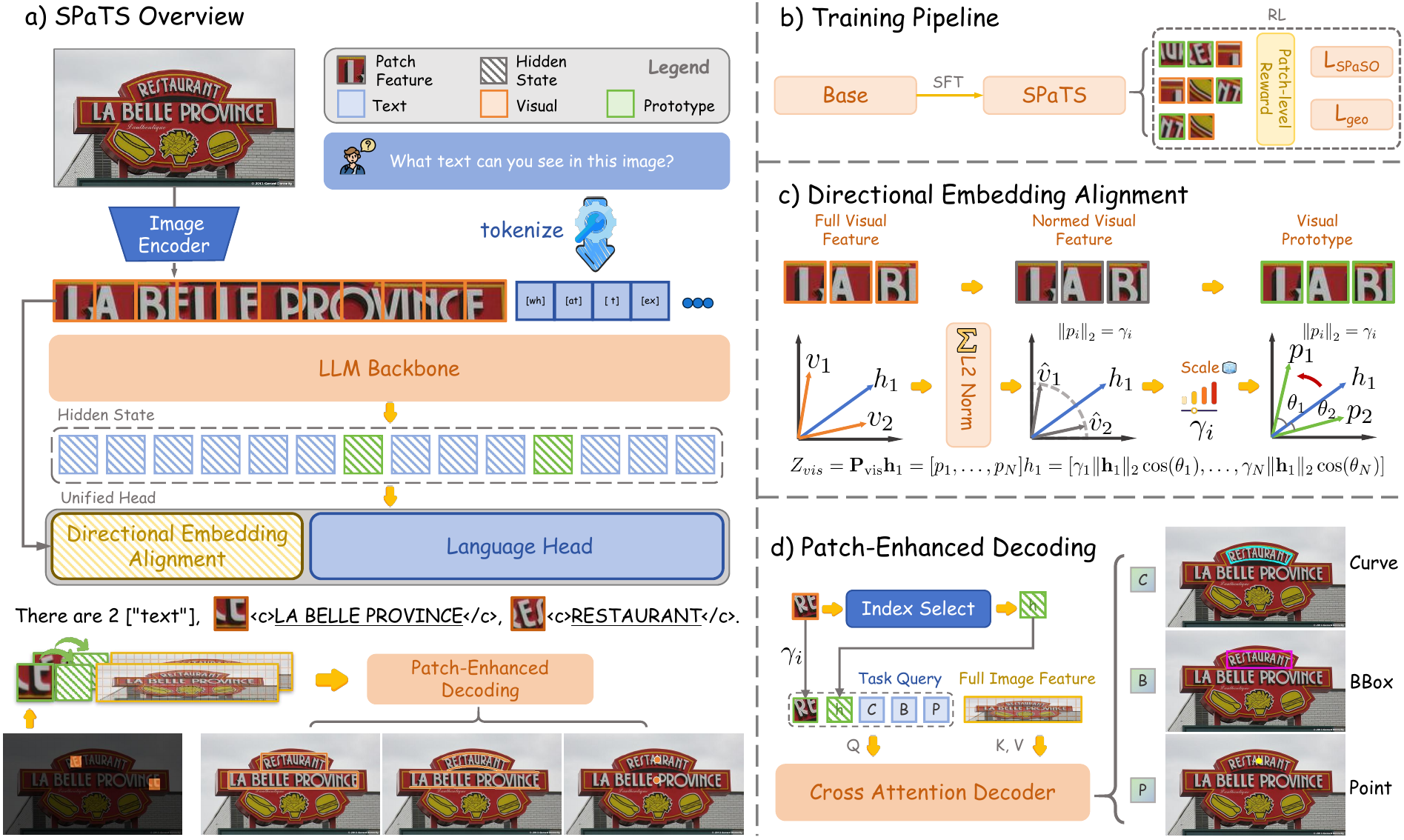}
  \vspace{-5pt}
  \caption{Overview of \textbf{SPaTS}. (a) Autoregressive spotting architecture. (b) Two-stage training with supervised fine-tuning and SPaSO. (c) DEA decouples feature direction and magnitude to form stable visual prototypes. (d) PED fuses the selected patch with language context for continuous geometry decoding.}
  \label{fig:overview}
  \vspace{-5pt}
\end{figure*}

\section{Related Work}

\subsection{Scene Text Spotting}
Scene text spotting has evolved from early end-to-end frameworks that jointly optimize detection and recognition \cite{liao2020real, liu2018fots} to more advanced methods for arbitrary-shaped text. Representative directions include segmentation-based approaches such as Mask TextSpotter~\cite{lyu2018mask}, regression-based methods such as ABCNet~\cite{liu2020abcnet} and TextDragon~\cite{feng2019textdragon}, and Transformer-based frameworks such as TESTR~\cite{zhang2022testr}, DeepSolo~\cite{ye2023deepsolo}, and SPTS v2~\cite{liu2023sptsv2}. More recently, MLLM-based OCR systems have reformulated spotting as a generative reasoning problem \cite{tu2024unicorns, li2024monkey, ocrgenbench2025zhang}. PaddleOCR-VL~\cite{cui2025paddleocrvl}, HunyuanOCR~\cite{hunyuanvisionteam2025hunyuanocr}, and dots.mocr~\cite{zheng2026multimodal} have achieved notable progress in spotting with MLLMs. Despite these advances, existing methods still face a trade-off between geometric precision and semantic reasoning ability.

\subsection{Unified Visual Tokenization for Grounding}
A related line of work studies how MLLMs bind language to visual evidence. Early methods such as Shikra~\cite{chen2023shikra} and Kosmos-2~\cite{peng2023kosmos} serialize grounding into coordinate or location tokens. Later approaches, including Groma~\cite{ma2024groma} and Osprey~\cite{yuan2024osprey}, replace coordinates with region-level proxies. More recent methods such as ClawMachine~\cite{ma2025clawmachine} and PaDT~\cite{su2026patch} move grounding closer to image-derived patch tokens. These advances motivate our use of visual tokens, but they are still not ideal for scene text spotting. In small, curved, and dense text scenes, coarse regions or multiple patches can mix the target text with nearby words and background content.

\subsection{Reinforcement Learning in MLLMs}
Reinforcement learning (RL)~\cite{ouyang2022training, guo2025deepseek} is used to align MLLMs with objectives beyond token-level likelihood, enabling models to optimize for reasoning quality, output structure, and performance. Group Relative Policy Optimization (GRPO)~\cite{shao2024deepseekmath} improves efficiency by removing the critic and normalizing rewards within groups of responses, thereby reducing memory and computation overhead. Works such as VLM-R1~\cite{shen2025vlm_r1}, VLM-RL~\cite{huang2024vlm_rl}, and Visual-RFT~\cite{liu2025visual_rft} extend this paradigm to multimodal reasoning and decision making, demonstrating RL's potential to improve perception\nobreak-\hspace{0pt}grounded outputs. However, most methods optimize textual spatial surrogates rather than geometric predictions, weakly coupling localization with policy objectives and limiting spatial rewards.

\begin{figure*}[t!]
  \centering
  \includegraphics[width=0.96\textwidth]{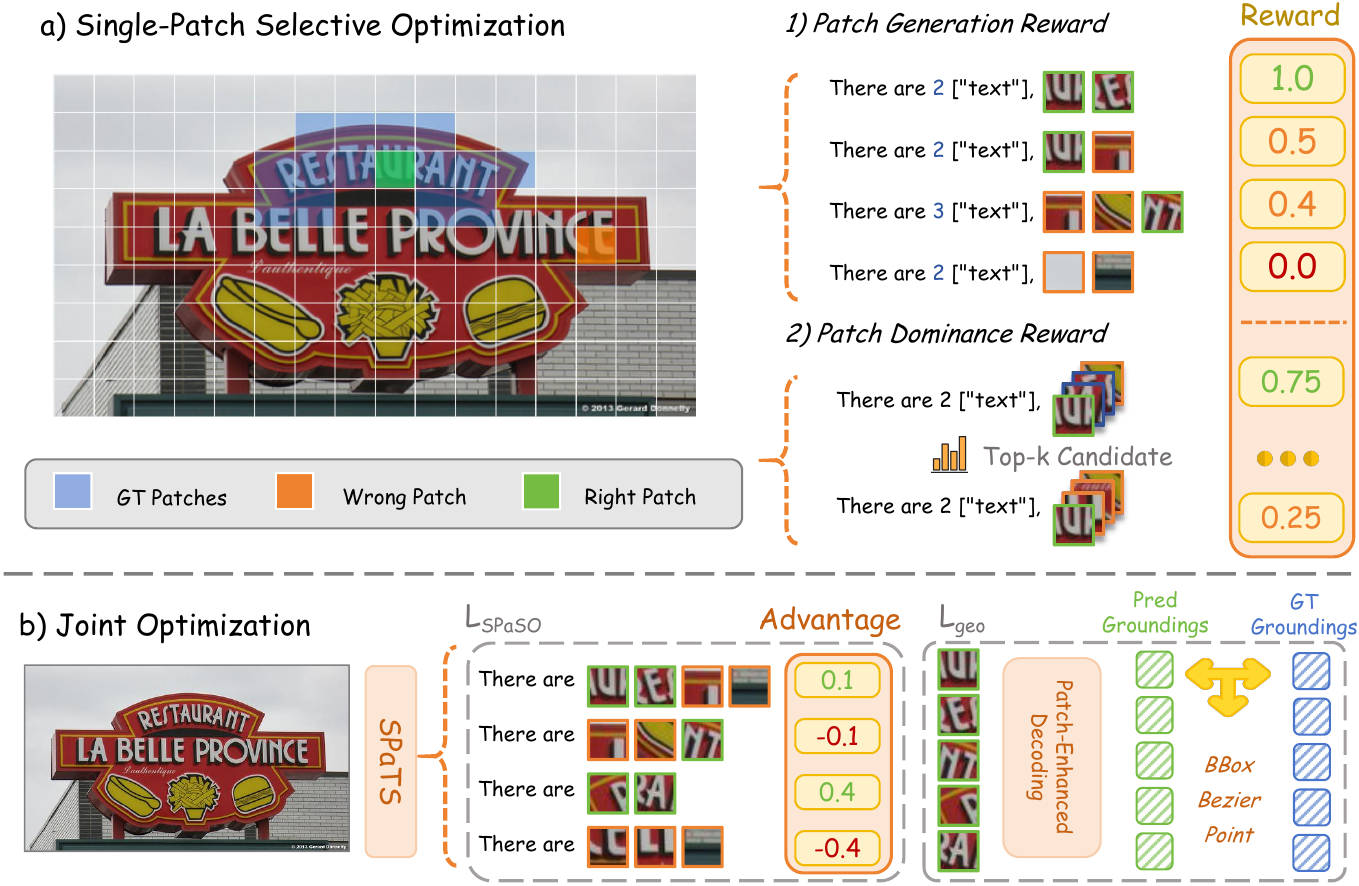}
  \caption{Illustration of \textbf{SPaSO}. (a) \textbf{Selection Optimization \& Reward Modeling:} The model selects a visual patch from candidates (blue: GT; orange: incorrect; green: correct). \textbf{Patch Generation Reward} measures spotting quality, while \textbf{Patch Dominance Reward} keeps correct patches among the top-k candidates. (b) \textbf{Joint Optimization:} SPaSO optimizes discrete selection with rewards while preserving geometry supervision for coordinate decoding.}
\Description{SPaSO framework: single-patch selection, generation and dominance rewards, and joint optimization with geometry supervision.}
  \label{fig:sparo}
\end{figure*}

\section{Method}

\subsection{Overall Architecture}
Following the pipeline design of PaDT~\cite{su2026patch}, SPaTS performs text recognition and spatial grounding within a vision-token framework.
As shown in Fig.~\ref{fig:overview}(a), given an input image $\mathcal{I} \in \mathbb{R}^{H \times W \times 3}$, where $H$ and $W$ denote the image height and width, the vision encoder $E(\cdot)$ extracts $N$ patch features
\begin{equation}
    \mathbf{V}_{\mathrm{patch}} = E(\mathcal{I}) = [\mathbf{v}_1, \ldots, \mathbf{v}_N] \in \mathbb{R}^{N \times d},
\end{equation}
where $\mathbf{v}_i \in \mathbb{R}^{d}$ is the visual feature of the $i$-th image patch and $d$ is the hidden dimension. Then all the current-image patches are adapted into image-specific visual prototypes:
\begin{equation}
    \mathbf{P}_{\mathrm{vis}} = f_{\mathrm{adapt}}(\mathbf{V}_{\mathrm{patch}}) = [\mathbf{p}_1, \ldots, \mathbf{p}_N] \in \mathbb{R}^{N \times d}.
\end{equation}
Here $\mathbf{p}_i \in \mathbb{R}^{d}$ is the visual prototype corresponding to the $i$-th patch. These prototypes are appended to both the text embedding table and the language head:
\begin{equation}
    \mathbf{E}_{\mathrm{uni}} = [\mathbf{E}_{\mathrm{text}};\mathbf{P}_{\mathrm{vis}}], \qquad
    \mathbf{W}_{\mathrm{uni}} = [\mathbf{W}_{\mathrm{text}};\mathbf{P}_{\mathrm{vis}}],
\end{equation}
where $\mathbf{E}_{\mathrm{text}}$ and $\mathbf{W}_{\mathrm{text}}$ denote the original textual embedding and language head weight, respectively. This yields an image-specific visual vocabulary $\mathcal{V}_{\mathrm{vis}} = \{v_1, \ldots, v_N\}$ derived from its native patch features, where token $v_i$ indexes visual prototype $\mathbf{p}_i$ for grounding.

Given a spotting prompt $\mathcal{P}$, the MLLM backbone performs autoregressive decoding over the unified vocabulary:
\begin{equation}
    \mathbf{h}_t = f_{\mathrm{mllm}}(\mathcal{P}, \mathcal{I}, y_{<t}), \qquad
    p(y_t \mid y_{<t}, \mathcal{P}, \mathcal{I}) = \mathrm{Softmax}(\mathbf{W}_{\mathrm{uni}}\mathbf{h}_t),
\end{equation}
where $\mathbf{h}_t \in \mathbb{R}^{d}$ is the hidden state at decoding step $t$, $y_{<t}$ denotes the previously generated tokens, and $y_t \in \mathcal{V}_{\mathrm{text}} \cup \mathcal{V}_{\mathrm{vis}}$ is the current output token. Text tokens continue the transcription, while corresponding visual patches are selected for grounding. In our setting, each text instance is associated with one visual token $v_{a_t}$, where $a_t$ is the selected patch index and $\mathbf{p}_{a_t}$ is its corresponding visual prototype:
\begin{equation}
    \mathbf{g}_t = f_{\mathrm{geo}}(\mathbf{h}_t, \mathbf{p}_{a_t}, \mathbf{V}_{\mathrm{patch}}),
\end{equation}
where $\mathbf{g}_t$ denotes the decoded geometry for the current instance, including bounding box, B\'ezier curve, or point sequence.

\subsection{Single-Patch Selective Optimization (SPaSO)}

\noindent \textbf{SFT Stage.} As shown in Figure~\ref{fig:overview}(b), before reinforcement learning, SPaTS is initialized through supervised fine-tuning (SFT) using ground-truth supervision. Given an image $\mathcal{I}$ and a spotting prompt $\mathcal{P}$, the MLLM generates the ground-truth interleaved sequence of text and visual patches, while PED decodes spatial grounding from the selected patch. The SFT loss consists of a token-level cross-entropy term and a direct geometry term:

\begin{equation}
  \mathcal{L}_{\mathrm{SFT}} = - \sum_t \log p(y_t^{\star} \mid y_{<t}^{\star}, \mathcal{P}, \mathcal{I}) + \mathcal{L}_{\mathrm{BBox}} + \mathcal{L}_{\mathrm{Bezier}} + \mathcal{L}_{\mathrm{Point}}.
\end{equation}
Here $y_t^{\star}$ denotes the ground-truth token at step $t$, and $y_{<t}^{\star}$ is the ground-truth prefix. The terms $\mathcal{L}_{\mathrm{BBox}}$, $\mathcal{L}_{\mathrm{Bezier}}$, and $\mathcal{L}_{\mathrm{Point}}$ represent the losses for bounding boxes, B\'ezier curves, and point sets, respectively, during supervised model training; their detailed mathematical formulations are provided in Appendix~\ref{sec:loss_formulation}.

\noindent \textbf{SPaSO Stage.} The main challenge of single-patch grounding lies in defining the \emph{optimal} patch, since no oracle label specifies the \emph{optimal} patch for the model. We formulate patch-selection as a discrete decision problem, where task-level feedback serves as the optimization signal. To address this, we propose Single-Patch Selective Optimization (SPaSO), a reinforcement learning objective built upon the GRPO framework.
As illustrated in Fig.~\ref{fig:sparo}(a), SPaSO applies RL to visual-token selection, while geometry regression is retained under matched continuous supervision. Following GRPO \cite{shao2024deepseekmath}, the policy $\pi_\theta$ is updated by maximizing the expected advantage over a sampled group of $G$ outputs:
\begin{equation}
\mathcal{J}_{\mathrm{SPaSO}}(\theta)
=
\mathbb{E}\!\left[
\frac{1}{G}\sum_{i=1}^{G}
\min\!\left(
\rho_i(\theta)\hat{A}_i,
\mathrm{clip}\!\left(\rho_i(\theta),1-\epsilon,1+\epsilon\right)\hat{A}_i
\right)
\right],
\end{equation}
where $\rho_i(\theta)=\pi_\theta(o_i\mid q)/\pi_{\theta_{\mathrm{old}}}(o_i\mid q)$ denotes the policy ratio and $\hat{A}_i$ is the normalized advantage. In SPaSO, the RL action space consists of sampling sequences characterized by patch indices rather than raw coordinate text. Consequently, the patch sampling distribution is optimized via the following patch-level rewards.

\noindent \textbf{Patch Generation Reward.} As shown in Fig.~\ref{fig:sparo}(a), we evaluate the quality of sampled patches through the lens of final spotting performance. A prediction is a \textbf{True Positive (TP)} if its corresponding patch resides within the ground-truth boundary and its generated text matches the ground truth within a predefined edit distance threshold. Combined with the average Intersection-over-Union ($\mathrm{IoU}_{\mathrm{avg}}$) of matched pairs, the reward $r_{\mathrm{gen}}$ is defined as:
\begin{equation}
r_{\mathrm{gen}} = \mathrm{HM}(F_1, \mathrm{IoU}_{\mathrm{avg}}) = \frac{2 \cdot F_1 \cdot \mathrm{IoU}_{\mathrm{avg}}}{F_1 + \mathrm{IoU}_{\mathrm{avg}}},
\end{equation}
where $F_1$ is the instance-level score calculated from the matching results. This formulation forces the policy to select patches that yield both high-fidelity transcriptions and precise geometric boundaries.

\noindent \textbf{Patch Dominance Reward.} To complement the sparse set-level reward, we further define a dense reward over the top-$k$ routed candidates. At each visual-token step $t$, let $p_t \in \mathbb{R}^{N}$ be the predictive distribution over candidate patches and let $\mathcal{G}_t$ be the set of ground-truth candidates. We measure ground-truth coverage within the Top-$k$ predictions:
\begin{equation}
r_{\mathrm{dom}} =
\frac{1}{T}\sum_{t=1}^{T}
\frac{1}{|\mathcal{G}_t|}
\sum_{g \in \mathcal{G}_t}
\mathbf{1}\!\left(g \in \mathrm{TopK}(p_t,k)\right).
\end{equation}
This reward encourages the policy to retain the correct patches before geometry decoding, providing sufficient optimization signals even when the generation reward is sparse or low.

\noindent \textbf{Joint Optimization.} As shown in Fig.~\ref{fig:sparo}(b), SPaSO is trained jointly with continuous geometry supervision. For each sampled pair $j$, PED decodes geometry as $\hat{\mathbf{g}}_j = f_{\phi}(\mathbf{F}^{(j)}_{\mathrm{ped}})$ and reuses the same bipartite matching $\mathcal{M}^{\star} = \operatorname{Match}(\hat{\mathcal{Y}}, \mathcal{Y})$ for geometry supervision. The geometry loss is
\begin{equation}
\begin{gathered}
\mathcal{L}_{\mathrm{geo}} =
\frac{1}{|\mathcal{M}^{\star}|}
\sum_{(j,m)\in\mathcal{M}^{\star}}
\left(
\mathcal{L}_{\mathrm{BBox}}^{j,m}
+\mathcal{L}_{\mathrm{Bezier}}^{j,m}
+\mathcal{L}_{\mathrm{Point}}^{j,m}
\right), \\
\mathcal{L}_{\mathrm{total}} =
-\mathcal{J}_{\mathrm{SPaSO}} + \mathcal{L}_{\mathrm{geo}}.
\end{gathered}
\end{equation}
Here $\mathcal{L}_{\mathrm{BBox}}$, $\mathcal{L}_{\mathrm{Bezier}}$, and $\mathcal{L}_{\mathrm{Point}}$ denote the regression losses for bounding boxes, B\'ezier curves, and point sets, respectively. In this way, SPaSO learns \emph{which} patch to select, while matched geometry supervision learns \emph{how} to recover its geometry.

\subsection{Directional Embedding Alignment (DEA)}
Since visual prototypes serve as decoding weights whose dot products with $\mathbf{h}_t$ determine the visual logits, the logits can be dominated by feature magnitude, biasing selection toward high-energy patches. To ensure the model learns precise \emph{directional} alignment, it is crucial to decouple feature direction from magnitude. Furthermore, to maintain the structural integrity of the pre-trained vision-language space, the visual prototypes $\mathbf{p}_i$ should remain aligned with the original image-patch features $\mathbf{v}_i$. We therefore adopt a design where the prototype is strictly proportional to the image feature, ensuring that the visual distribution of the unified head preserves image-specific semantics.

As shown in Fig.~\ref{fig:overview}(c), DEA implements this by parameterizing each visual prototype via a normalized direction and a learnable scale. Given patch features $\mathbf{V}_{\mathrm{patch}}=[\mathbf{v}_1,\ldots,\mathbf{v}_N]$, we compute:
\begin{equation}
\hat{\mathbf{v}}_i = \frac{\mathbf{v}_i}{\|\mathbf{v}_i\|_2}, \qquad
\gamma_i = \mathbf{W}_{\gamma}\mathbf{v}_i, \qquad
\mathbf{p}_i = \gamma_i \hat{\mathbf{v}}_i.
\end{equation}
The resulting prototypes are stacked as $\mathbf{P}_{\mathrm{vis}}=[\mathbf{p}_1,\ldots,\mathbf{p}_N]$ and appended to the textual head, yielding the unified output matrix $\mathbf{W}_{\mathrm{uni}} = [\mathbf{W}_{\mathrm{text}};\mathbf{P}_{\mathrm{vis}}]$.

Under this decoupled parameterization, the visual logits in the unified head take the form:
\begin{equation}
\begin{aligned}
\mathbf{Z} 
=
[\mathbf{W}_{\mathrm{text}}\mathbf{h}_t;\gamma_1\|\mathbf{h}_t\|_2\cos(\theta_1), \ldots, \gamma_N\|\mathbf{h}_t\|_2\cos(\theta_N)],
\end{aligned}
\end{equation}
where $\theta_i$ is the angle between $\mathbf{h}_t$ and $\hat{\mathbf{v}}_i$. Isolating the scale $\gamma_i$ prevents raw feature norms from dominating selection decisions. Crucially, by freezing $\gamma_i$ after warmup, DEA forces the model to focus on learning directional semantic alignment (cosine similarity).

At a fixed decoding step, $\|\mathbf{h}_t\|_2$ is shared across all candidate patches, and the cross-entropy over visual logits becomes:
\begin{equation}
\mathcal{L}_{\mathrm{CE}}
=
-\log
\frac{\exp\!\left(\gamma_{\mathrm{target}}\|\mathbf{h}_t\|_2\cos(\theta_{\mathrm{target}})\right)}
{\sum_j \exp(z_j)}.
\end{equation}

\subsection{Patch-Enhanced Decoding (PED)}

After SPaTS selects the visual token $\mathbf{v}_{a_t}$ and its corresponding prototype $\mathbf{p}_{a_t}$, the model still needs to recover $\mathbf{g}_t = f_{\mathrm{geo}}(\mathbf{h}_t,\mathbf{p}_{a_t},\mathbf{V}_{\mathrm{patch}})$. Using only $\mathbf{p}_{a_t}$ is insufficient: it indicates where to attend, but does not fully indicate which text instance is being decoded or how its boundary should be traced. PED addresses this by combining the routed prototype with the language hidden state and using the fused query to read from the full-image feature map.

Specifically, as shown in Fig.~\ref{fig:overview}(d), given $\mathbf{p}_{a_t}$ and $\mathbf{h}_t$, we form an object query
\begin{equation}
    \mathbf{q}_t = \phi\!\left([\mathbf{p}_{a_t} \,||\, \mathbf{h}_t]\right),
\end{equation}
where $\mathbf{p}_{a_t}$ provides local visual evidence and $\mathbf{h}_t$ specifies the decoded text content. We then prepend four learnable readout tokens for box, point, curve, and score prediction:
\begin{equation}
    \mathbf{Q}_t^0 = [\mathbf{q}_t; \mathbf{t}_{\mathrm{bbox}};\mathbf{t}_{\mathrm{point}};\mathbf{t}_{\mathrm{curve}};\mathbf{t}_{\mathrm{score}}],
    \qquad
    \mathbf{K}_t = \mathbf{V}_t = \mathbf{V}_{\mathrm{patch}}.
\end{equation}
The query sequence $\mathbf{Q}_t^0$ is refined by a three-layer cross-attention decoder over $\mathbf{V}_{\mathrm{patch}}$, and the first four output tokens are fed to the bounding-box, point, B\'ezier-curve, and score heads to predict $\mathbf{g}_t$.

\begin{table*}[t]
  \caption{Comparison of text spotting performance. \textbf{SPaTS-2B} and \textbf{SPaTS-4B} denote the 2B and 4B model variants, respectively. "P", "R", and "F" represent Precision, Recall, and F-measure. "E2E" denotes end-to-end spotting results under "None", "Full" (Total-Text/CTW1500), and "S", "W", "G" (ICDAR 2015) lexicon protocols. Bold and underlined values indicate the best and second-best results.}
  \label{tab:main_result}
  \resizebox{\linewidth}{!}{
  \begin{tabular}{lcccccccccccccccc}
  \toprule
  \multirow{3}{*}{Methods}             & \multicolumn{5}{c}{Total-Text}   & \multicolumn{5}{c}{CTW1500}      & \multicolumn{6}{c}{ICDAR 2015}          \\ \cmidrule(lr){2-6} \cmidrule(lr){7-11} \cmidrule(lr){12-17}
  & \multicolumn{3}{c}{Detection}   & \multicolumn{2}{c}{E2E}    & \multicolumn{3}{c}{Detection}   & \multicolumn{2}{c}{E2E} & \multicolumn{3}{c}{Detection}   & \multicolumn{3}{c}{E2E}          \\
  \cmidrule(lr){2-4} \cmidrule(lr){5-6} \cmidrule(lr){7-9}  \cmidrule(lr){10-11} \cmidrule(lr){12-14} \cmidrule(lr){15-17} 
                      & P    & R    & F    & None & Full & P    & R    & F    & None & Full & P    & R    & F    & S    & W    & G  
                      \\ \midrule
  \mainresultgroup{Closed-Source}
  \midrule
  Claude-Sonnet-4.6~\cite{anthropic2026claudesonnet46}            & 43.1  & 15.5  & 22.8 & 14.1 & 15.9 &  64.7  & 46.8   & 54.3 & 30.9 &  35.3  & 20.6 & 10.1 & 13.5 & 11.7 & 10.3  & 10.2  \\
  Gemini-3.1-Flash~\cite{googleteam2025gemini3flash}            & 68.9  &  59.2 & 63.7 & 56.6 & 59.8 &  47.1   & 68.5   &  55.8   & 43.3 & 48.5  & 48.9 & 38.8 & 43.2 & 38.0 & 36.9 & 36.5 \\
  GPT-5.4~\cite{openai2026gpt54}      & 53.4  & 19.5  & 28.6 & 17.2 & 20.2 &  44.3   & 34.2   &  38.6   & 27.9 & 32.4  & 37.9 & 18.2 & 24.6 & 12.4 & 11.9 & 11.9 \\
  seed-2.0-pro~\cite{bytedance2026seed2.0}            & 69.8  & 25.8  & 37.6 & 30.8 & 35.6 &  81.7   & 78.5   & 80.1   & 62.8 & 76.5  & 37.1 & 26.9 & 31.2 & 27.6 & 27.0 & 26.9 \\
  \midrule
  \mainresultgroup{Open-Source}
  \midrule
  Kimi-K2.5~\cite{moonshotai2026kimik25}            & 73.0  & 35.8  & 48.1 & 42.3 & 46.3 & 72.3  & 77.8 & 75.0 & \ranksecond{58.2} & 70.1  & 48.5 & 36.9 & 41.9 & 35.9 & 35.1  & 34.6  \\
  Qwen3.5-397B-A17B~\cite{qwenteam2026qwen35}            & 62.7  & 34.5  & 44.5 & 38.2 & 42.5 &  45.5   & 51.3   & 48.2    & 37.1 & 44.6  & 55.8 & 42.8 & 48.4 & 42.6 & 41.9  &  41.3 \\
  InternVL3.5-241B-A28B~\cite{wang2025internvl3_5}     & 31.6  & 9.0  & 14.1 & 9.2 & 10.2 &  58.1   & 52.5   & 55.1   & 29.3 & 35.9  & 19.5 & 11.1 & 14.1 & 7.4 & 7.1 & 6.9 \\
  GLM-4.6V~\cite{zai2025glm46v}          &  65.7  & 18.2  & 28.6 & 23.5 & 27.1 &  68.0  &  54.3  &  60.4  & 47.2 & 56.0  & 38.9 & 21.3 & 27.6 & 23.8 & 22.8 & 22.6 \\
  \midrule
  \mainresultgroup{Specialized}
  \midrule
  HunyuanOCR~\cite{hunyuanvisionteam2025hunyuanocr}      & 77.9  & 35.9  & 49.2 & 41.5 & 44.3 &  76.5   & \rankfirst{89.9}   &  \rankfirst{82.7}   & \rankfirst{63.8} & \ranksecond{77.7}  & 56.8 & 49.5 & 52.9 & 43.2 & 40.1 & 39.8  \\
  PaddleOCR-VL-1.5~\cite{cui2025paddleocrvl}      & 69.2  & 36.8 & 48.1 & 37.8 & 45.1 &  62.9   & \ranksecond{80.5}  &  70.6   & 50.4 & 66.2  & 42.6 & 59.8 & 49.8 & 45.1 & 44.4 & 42.1 \\
  dots.mocr~\cite{zheng2026multimodal}      & 60.3  & 36.9  & 45.8 & 39.3 & 44.1 &  29.4   & 46.1   &  35.9   & 26.6 & 33.8  & 7.3 & 29.4 & 11.7 & 10.8 & 10.7 & 10.7  \\
  \midrule
  \mainresultgroup{Ours}
  \midrule
  SPaTS-2B               & \rankfirst{82.3} & \ranksecond{80.9} & \rankfirst{81.7} & \ranksecond{63.3} & \ranksecond{74.3} & \ranksecond{87.9} & 75.8 & 81.4 & 47.0 & 75.1 & \rankfirst{89.1} & \ranksecond{70.0} & \ranksecond{78.4} & \ranksecond{72.1} & \ranksecond{69.1} & \ranksecond{61.6} \\
  SPaTS-4B               & \ranksecond{80.4} & \rankfirst{82.7} & \ranksecond{81.5} & \rankfirst{64.9} & \rankfirst{74.9} & \rankfirst{89.9} & 75.9 & \ranksecond{82.4} & 51.2 & \rankfirst{79.6} & \ranksecond{86.2} & \rankfirst{73.9} & \rankfirst{79.5} & \rankfirst{73.4} & \rankfirst{71.7} & \rankfirst{65.7} \\
  \bottomrule
  \end{tabular}
  }
  \end{table*}

\section{Experiments}
\subsection{Experimental Setup}
\noindent \textbf{Datasets.} We train SPaTS on a multi-source collection for geometric robustness, including Curved Synthetic 150k~\cite{liu2020abcnet} for pre-training, supplemented by real-world datasets: MLT-2017~\cite{nayef2017icdar}, ICDAR 2013/2015~\cite{karatzas2013icdar,karatzas2015icdar}, Total-Text~\cite{chng2017total}, and SCUT-CTW1500~\cite{liu2019curved}. These cover diverse challenges including horizontal, multi-oriented, and highly curved text. Evaluation is performed on the test sets of Total-Text, SCUT-CTW1500, and ICDAR 2015.

\noindent \textbf{Implementation Details.}
We build SPaTS on Qwen3-VL 2B/4B~\cite{bai2025qwen3vl}, incorporating a three-layer cross-attention decoder for fine-grained geometric refinement~\cite{su2026patch}. The training follows a two-stage pipeline: (1) \textbf{SFT stage}: The model is trained for 16 epochs using the AdamW optimizer with a learning rate of $4\times10^{-5}$ and a cosine decay schedule. Training is executed on 8 NVIDIA A6000 GPUs with an effective batch size of 128. (2) \textbf{RL stage}: We further optimize visual patch selection via SPaSO for 30 epochs using GRPO~\cite{shao2024deepseekmath} with a group size of 4 and a reduced learning rate of $5\times10^{-6}$. The vision encoder and scale head are both frozen during this phase. Additional implementation details are provided in Appendix~\ref{sec:training_configuration}.

\begin{table}[t]

\centering
\caption{End-to-end results of grounding formulations using the same backbone after SFT. ``Coord-Text'' denotes coordinate generation in text space, while ``Multi-Patch'' and ``Single-Patch'' denote visual-token grounding with multiple routed patches and a single routed patch, respectively. ``P'', ``R'', and ``F'' denote Precision, Recall, and F-measure.}
\vspace{+5pt}
\label{tab:comparison}
\resizebox{\linewidth}{!}{
\begin{tabular}{l ccc ccc ccc}
\toprule
\multirow{2}{*}{Methods} 
& \multicolumn{3}{c}{Total-Text}
& \multicolumn{3}{c}{CTW1500}
& \multicolumn{3}{c}{ICDAR 2015} \\
\cmidrule(lr){2-4}
\cmidrule(lr){5-7}
\cmidrule(lr){8-10}
& P & R & F
& P & R & F
& P & R & F \\
\midrule
Coord-Text & \ranksecond{53.2}  & \ranksecond{46.3}  & \ranksecond{49.5} & \ranksecond{39.7} & \ranksecond{45.8} &  \ranksecond{42.6}   &  \ranksecond{48.5}  & 39.1    & 43.3 \\
Multi-Patch & 30.6  & 16.7  & 21.6 & 34.7 & 24.3 &  28.6   &  47.8  & \ranksecond{39.8}    & \ranksecond{43.4} \\
Single-Patch & \rankfirst{63.0}  & \rankfirst{64.5}  & \rankfirst{63.7} & \rankfirst{40.3} & \rankfirst{57.9} &  \rankfirst{47.6}   &  \rankfirst{66.1}  & \rankfirst{57.3}    & \rankfirst{61.4} \\
\bottomrule
\end{tabular}
}
\vspace{-5pt}
\end{table}

\begin{table}[t]
\tiny
\centering
\vspace{-5pt}
\caption{End-to-end results across post-training stages. "Base" is the pre-trained model; "SFT" and "SPaSO" denote subsequent supervised fine-tuning and reinforcement learning on target datasets, respectively.}
\vspace{-5pt}
\label{tab:training_evolution}
\renewcommand{\arraystretch}{0.83}
\resizebox{0.7\linewidth}{!}{
\begin{tabular}{l cc cc}
\toprule[0.5pt]
\multirow{2}{*}{Stage} 
& \multicolumn{2}{c}{Total-Text}
& \multicolumn{2}{c}{CTW1500} \\
\cmidrule(lr){2-3}
\cmidrule(lr){4-5}
& None & Full
& None & Full \\
\midrule[0.3pt]
Base
& 56.8 & 63.7
& 41.8 & 47.6 \\
\midrule[0.3pt]
SFT
& 61.2 & 69.7
& \rankfirst{53.1} & 58.9 \\
SPaSO
& \rankfirst{64.9} & \rankfirst{74.9}
& 51.2 & \rankfirst{79.6} \\
\bottomrule[0.5pt]
\end{tabular}
}
\end{table}

\subsection{Main Results}

\noindent \textbf{Comprehensive State-of-the-Art Performance.} 
As shown in Table~\ref{tab:main_result}, both SPaTS variants (2B and 4B) achieve significant improvements across three benchmarks: Total-Text, CTW1500, and ICDAR 2015. They dominate almost all core metrics in both text detection and end-to-end (E2E) spotting, demonstrating robust generalization and superior capability across diverse text scenarios.

\noindent \textbf{Superiority over Specialized Models.} 
Compared with specialized OCR models such as HunyuanOCR and PaddleOCR-VL-1.5, SPaTS demonstrates performance gains, particularly in the challenging E2E spotting task. On the ICDAR 2015 benchmark, for instance, SPaTS raises the E2E metrics (S/W/G) from previous state-of-the-art levels of approximately 50\%. This notable improvement in accuracy suggests that our approach can effectively address complex spotting scenarios where traditional methods have faced limitations.

\noindent \textbf{High Parameter Efficiency vs. MLLMs.} 
Despite utilizing a relatively lightweight 2B/4B architecture, SPaTS decisively outperforms both massive open-source MoE models and frontier closed-source systems (e.g., GPT-5.4, Claude-Sonnet-4.6, and Gemini-3.1-Flash), achieving superior results with significantly higher parameter efficiency. This highlights the extreme efficiency of our architectural design for vision-language alignment, effectively achieving superior text spotting results with a fraction of the parameter scale.

\begin{table}[t]
\scriptsize
\centering
\caption{End-to-end results of different localization representations. "Bezier", "BBox", and "Point" denote explicit coordinate-based variants using Bezier curves, bounding boxes, and central points, respectively, while "Patch" represents our single-patch selection formulation.}
\vspace{-5pt}
\label{tab:coordinate_ablation}
\renewcommand{\arraystretch}{0.8}
\resizebox{0.8\linewidth}{!}{
\begin{tabular}{l ccc ccc}
\toprule[0.5pt]
\multirow{2}{*}{Variants}
& \multicolumn{3}{c}{Total-Text}
& \multicolumn{3}{c}{CTW1500} \\
\cmidrule(lr){2-4}
\cmidrule(lr){5-7}
& P & R & F
& P & R & F \\
\midrule[0.3pt]

Bezier
& 53.7 & 52.8 & 53.3
& 60.2 & 50.9 & 55.1\\

BBox
& 73.8 & \ranksecond{75.9} & 74.9
& 87.1 & 73.5 & 79.7\\

Point
& \rankfirst{75.4} & 75.6 & \rankfirst{75.5}
& \ranksecond{87.6} & \ranksecond{73.9} & \ranksecond{80.1} \\
\midrule[0.3pt]
Patch
& \ranksecond{73.9} & \rankfirst{76.4} & \ranksecond{75.2}
& \rankfirst{88.5} & \rankfirst{74.9} & \rankfirst{81.2} \\

\bottomrule[0.5pt]
\end{tabular}
}
\vspace{-5pt}
\end{table}

\begin{table}[t]
\centering
\caption{\textbf{Ablation study on SPaTS components.} $L_2$ norm and Scale denote $L_2$ normalization and learnable scaling, respectively; $\mathbf{h}_t$ and $\mathbf{p}_{a_t}$ represent the language hidden state and routed prototype. Results are reported on the SFT-trained model to isolate the effects of different training stages.}
\label{tab:module_ablation}
\resizebox{\linewidth}{!}{
\begin{tabular}{cc ccc ccc ccc}
\toprule

\multicolumn{2}{c}{\multirow{2}{*}{Components}}
& \multicolumn{3}{c}{Total-Text}
& \multicolumn{3}{c}{CTW1500}
& \multicolumn{3}{c}{ICDAR 2015} \\
\cmidrule(lr){3-5} \cmidrule(lr){6-8} \cmidrule(lr){9-11}
&  
& P & R & F
& P & R & F
& P & R & F \\
\midrule
\rowcolor{black!8} $L_2$ Norm &  \multicolumn{1}{c|}{Scale} & \multicolumn{9}{c}{\textit{DEA}} \\
\midrule

\cellcolor{lightred}\xmark & \cellcolor{lightred}\xmark
& 53.1 & 53.7 & 53.4
& 29.6 & 26.9 & 28.2
& 41.7 & 47.8 & 44.5 \\

 \cellcolor{lightgreen}\cmark & \cellcolor{lightred}\xmark
& 60.3 & \ranksecond{61.7} & \ranksecond{61.0}
& \ranksecond{32.6} & 54.7 & \ranksecond{40.9}
& 53.5 & 44.6 & 48.7 \\

 \cellcolor{lightred}\xmark & \cellcolor{lightgreen}\cmark
& \rankfirst{68.5} & 47.0 & 55.8
& 31.1 & \rankfirst{58.4} & 40.6
& \ranksecond{63.8} & \ranksecond{53.2} & \ranksecond{58.0} \\

\cellcolor{lightgreen}\cmark & \cellcolor{lightgreen}\cmark
& \ranksecond{63.0} & \rankfirst{64.5} & \rankfirst{63.7}
& \rankfirst{40.3} & \ranksecond{57.9} & \rankfirst{47.6}
& \rankfirst{66.1} & \rankfirst{57.3} & \rankfirst{61.4} \\
\midrule
\rowcolor{black!8} $\mathbf{h}_t$ &  \multicolumn{1}{c|}{$\mathbf{p}_{a_t}$} & \multicolumn{9}{c}{\textit{PED}} \\
\midrule

 \cellcolor{lightgreen}\cmark & \cellcolor{lightred}\xmark
& 50.9 & 35.1 & 41.5
& 31.9 & 32.3 & 32.1
& 43.0 & 30.4 & 35.6 \\

\cellcolor{lightgreen}\cmark & \cellcolor{lightgreen}\cmark
& \rankfirst{63.0} & \rankfirst{64.5} & \rankfirst{63.7}
& \rankfirst{40.3} & \rankfirst{57.9} & \rankfirst{47.6}
& \rankfirst{66.1} & \rankfirst{57.3} & \rankfirst{61.4} \\

\bottomrule
\end{tabular}
}
\vspace{-10pt}
\end{table}

\subsection{Further Analysis}

\noindent \textbf{Superiority of Single-Patch Grounding.} 
As evidenced in Table~\ref{tab:comparison}, \emph{Single-Patch} selection consistently outperforms both \emph{Coord-Text} and \emph{Multi-Patch} formulations. Unlike text-space tokenization which suffers from precision loss, our approach establishes a direct semantic-to-visual correspondence. The poor performance of \emph{Multi-Patch} further suggests that multiple visual tokens introduce substantial selection ambiguity; thus, by constraining SPaTS to a single high-quality patch, we achieve a more efficient and stable grounding interface with superior representational sparsity.

\noindent \textbf{Synergy between SFT and RL Stages.} 
The performance evolution in Table~\ref{tab:training_evolution} highlights the distinct optimization goals of each post-training stage. While \textbf{SFT} maintains a slight lead in blind settings by strictly minimizing character-level cross-entropy, \textbf{SPaSO} leverages edit-distance tolerance to prioritize semantic utility. This strategic shift facilitates a massive 20.7\% leap under the \emph{Full} lexicon protocol, confirming that RL-optimized selection logic is significantly more effective at aligning visual prototypes with candidate lexical sequences than standard supervised fine-tuning.

\noindent \textbf{Representational Robustness and Complexity.}
\noindent Beyond specific training stages, SPaTS exhibits remarkable robustness across diverse grounding protocols, maintaining consistent performance as evidenced in Table~\ref{tab:coordinate_ablation}. Notably, our framework maintains stable performance even with \emph{Bezier} curves—a highly non-linear parameterization nearly impossible to learn within standard text-space paradigms. While \emph{BBox} excels in blind localization, \emph{Point} and \emph{Patch} show superior scalability, with \emph{Patch} reaching an impressive F-measure of 81.2\% on CTW1500. This underscores that discrete visual patches serve as more robust semantic anchors for selection and alignment than discrete geometric text coordinates. More detailed analyses are provided in Appendix~\ref{sec:extended_analysis}.

\begin{table}[t]
  \centering
  \caption{\textbf{Ablation of Reward Modeling in SPaSO.} "Gen" and "Dom" denote Patch Generation Reward and Patch Dominance Reward, respectively. End-to-end Performance is reported across three text spotting benchmarks. Bold and underlined values indicate the best and second-best results.}
  \label{tab:reward_ablation}
  \resizebox{\linewidth}{!}{
  \begin{tabular}{cc ccc ccc ccc}
  \toprule

  \multicolumn{2}{c|}{Reward}
  & \multicolumn{3}{c}{Total-Text}
  & \multicolumn{3}{c}{CTW1500}
  & \multicolumn{3}{c}{ICDAR 2015} \\
  \cmidrule(lr){1-2}
  \cmidrule(lr){3-5}
  \cmidrule(lr){6-8}
  \cmidrule(lr){9-11}

  Gen & \multicolumn{1}{c|}{Dom} & P & R & F
  & P & R & F
  & P & R & F \\
  \midrule

  \cellcolor{lightgreen}\cmark & \cellcolor{lightred}\xmark
  & \ranksecond{72.8} & \ranksecond{74.4} & \ranksecond{73.6}
  & \ranksecond{81.4} & \ranksecond{68.6} & \ranksecond{74.5}
  & \ranksecond{76.6} & 59.3 & \ranksecond{66.8} \\

   \cellcolor{lightred}\xmark & \cellcolor{lightgreen}\cmark
  & 66.7 & 67.8 & 67.3
  & 58.9 & 57.1 & 58.0
  & 65.5 & \ranksecond{62.5} & 64.0 \\

  \cellcolor{lightgreen}\cmark & \cellcolor{lightgreen}\cmark
  & \rankfirst{73.8} & \rankfirst{75.9} & \rankfirst{74.9}
  & \rankfirst{86.1} & \rankfirst{74.2} & \rankfirst{79.7}
  & \rankfirst{82.1} & \rankfirst{64.0} & \rankfirst{72.0} \\

  \bottomrule[0.5pt]
  \end{tabular}
  }
\end{table}

\subsection{Ablation Study}

\noindent \textbf{Contribution of Architectural Components.}
As evidenced in Table~\ref{tab:module_ablation}, the \textbf{DEA} module relies on the joint application of $L_2$ normalization and learnable scaling to ensure robust similarity computation. 
Without this stabilization, selection is dominated by feature norms rather than semantic relevance, causing detrimental feature drift. 
Critically, the \textbf{PED} module acts as a necessary bottleneck for visual grounding; while the language state $\mathbf{h}_t$ maintains temporal coherence, it struggles with fine-grained occlusions. 
The collapse in accuracy following the removal of the corresponding prototype $\mathbf{p}_{a_t}$ confirms that grounding the language prior in explicit visual signals is essential to prevent model hallucination.

\noindent \textbf{Dual Reward Synergy.}
Table~\ref{tab:reward_ablation} highlights the complementarity of our reward design in \textbf{SPaSO}. The Patch Generation Reward ($r_{\mathrm{gen}}$) serves as the primary objective, ensuring that selection decisions directly optimize end-to-end spotting. However, the Patch Dominance Reward ($r_{\mathrm{dom}}$) is essential for stabilization—it provides dense credit assignment when $r_{\mathrm{gen}}$ is sparse, ensuring the model identifies the most informative patch even in early RL stages. Combined, they yield the most robust selection policy across all benchmarks.

\begin{figure}[t]
  \centering
  \includegraphics[width=\linewidth]{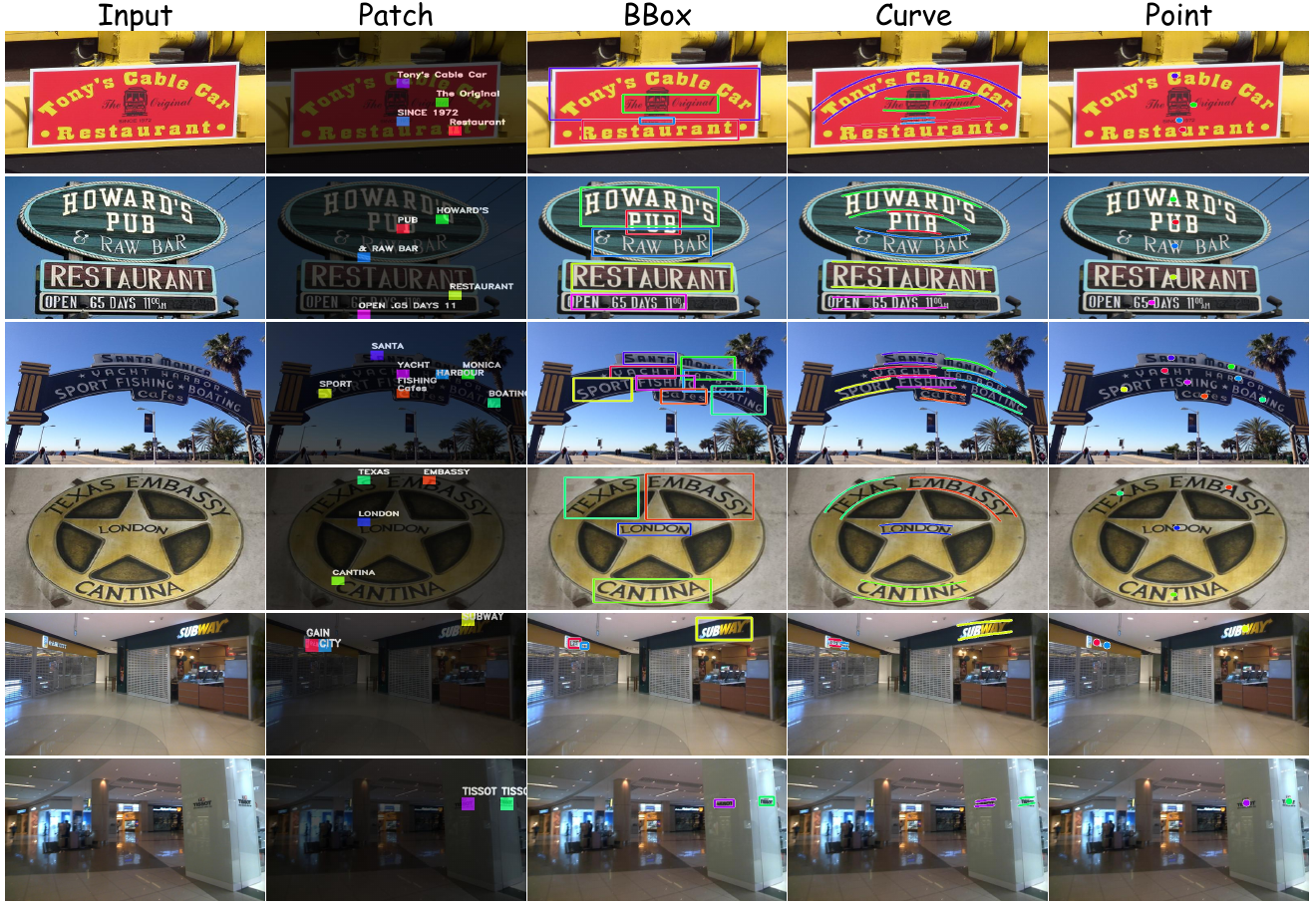}
  \caption{\textbf{Qualitative comparison of localization representations.} Columns show the input, SPaTS-routed patches and labels, and BBox/Curve/Point results (green: GT; red: prediction). Our method handles curved text (rows 1--2, CTW1500), dense or multi-oriented text (rows 3--4, Total-Text), and low-contrast text (rows 5--6, ICDAR 2015). Zoom in for better view.}
  \vspace{-10pt}
  \label{fig:case_study}
\end{figure}

\subsection{Qualitative Evaluation}

\noindent \textbf{Robustness Across Complex Scenarios.} 
As illustrated in Fig.~\ref{fig:case_study}, SPaTS effectively navigates curved and irregular geometries that often hinder conventional VLMs. 
On CTW1500 and Total-Text, it maintains precise instance-level grounding where BBox and Curve baselines suffer from geometric drift. 
By employing selected patches as stable semantic anchors, SPaTS ensures transcriptions remain strictly coupled to instances despite non-rigid boundaries. 
This robustness extends to ICDAR 2015, where it captures weak or occluded text frequently overlooked by general-purpose MLLMs.

\begin{figure}[t]
  \centering
  \includegraphics[width=0.97\linewidth]{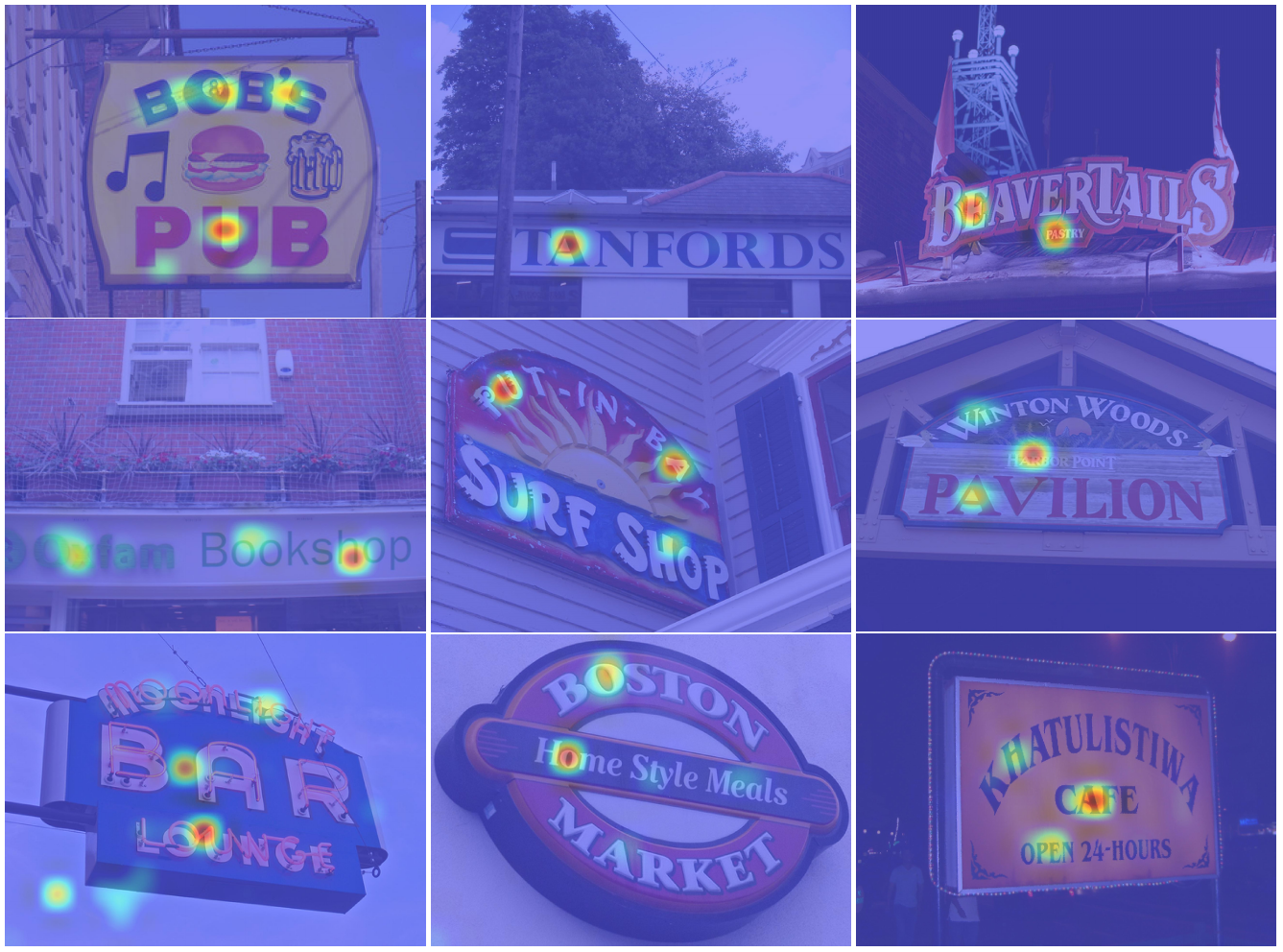}
  \caption{Attention heat map during visual-token selection. Brighter regions indicate stronger focus on the grounded text instance. Zoom in for better view.}
  \Description{Attention heatmaps concentrated on the correct text regions during visual-token selection.}
  \label{fig:attention}
\end{figure}

\noindent \textbf{Interpretability of Visual Selection.} 
We examine the underlying selection mechanism through the cross-attention heatmaps visualized in Fig.~\ref{fig:attention}. 
During the visual-token selection phase, attention is observed to concentrate sharply on the target text regions, effectively isolating the relevant features from surrounding clutter and distractors. 
Such a focused pattern explains the reliability of SPaTS in crowded scenes; the high degree of alignment between attention peaks and text geometry confirms that our lightweight interface successfully distills the most informative local evidence while suppressing interference in downstream decoding.

\begin{figure}[t]
  \centering
  \includegraphics[width=0.97\linewidth]{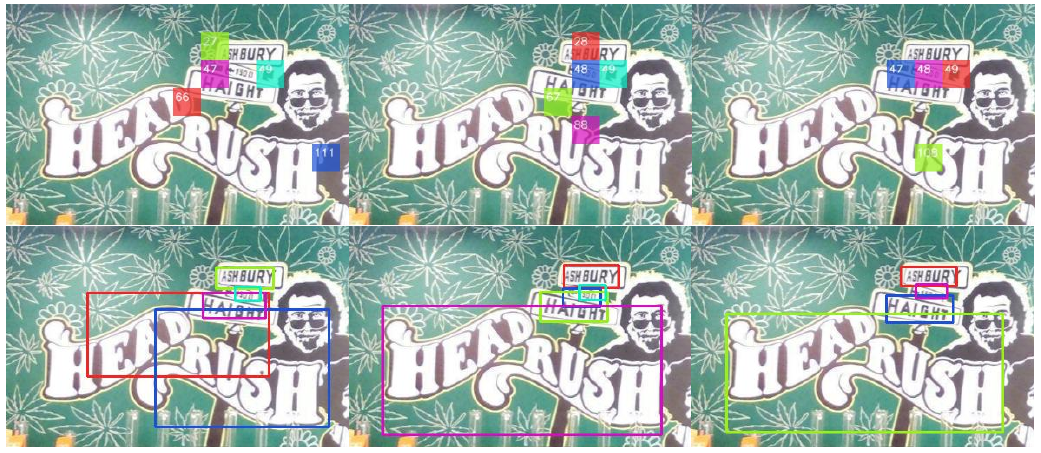}
  \caption{Localization quality varies across different sampled patches. Different selection choices can lead to noticeably different grounding quality, even for the same text instance. Zoom in for better view.}
  \Description{Examples showing that sampling different image patches leads to different localization quality for the same text instance.}
  \label{fig:rl_necessity}
\end{figure}

\noindent \textbf{Impact of Selection on Spotting Quality.} 
The correlation between selection precision and final spotting fidelity is clearly demonstrated in Fig.~\ref{fig:rl_necessity}. 
Our analysis reveals that localization quality is fundamentally sensitive to selection decisions at inference: while an optimal, centered patch yields tight and accurate results, peripheral or mismatched patches lead to immediate grounding degradation. 
These observations validate the core motivation of SPaSO---that optimizing discrete patch selection is a prerequisite for high-precision spotting in practice, as the decoder's performance is inherently upper-bounded by the quality of the visual evidence it receives.

\section{Limitations}

Despite its effectiveness, SPaTS still has several limitations. First, although SPaSO optimizes selection without oracle labels, the RL stage introduces additional training overhead and depends on carefully designed reward shaping for stable optimization. Second, although DEA stabilizes prototypes and improves alignment, freezing the magnitude branch after warmup may reduce flexibility under extreme illumination changes, where feature scales can vary more drastically than in standard settings. These factors may limit the model's robustness and efficiency in highly challenging real-world scenarios. Future work will focus on reducing the optimization cost of SPaSO and improving robustness under more extreme conditions.

\section{Conclusion}

In this paper, we introduce \textbf{SPaTS}, a new framework that explicitly rethinks visual grounding granularity for MLLM-based scene text spotting. By routing each text instance through a single, instance-specific, task-optimized anchor visual token and recovering geometry via full-image refinement, we replace noisy and redundant multi-patch aggregation with a more compact, precise formulation that enforces a tighter coupling between visual evidence and geometric prediction. We further propose \textbf{SPaSO}, a reinforcement learning method that reformulates the latent visual token assignment problem as a discrete policy optimization problem without extra annotations, bypassing the need for oracle patch supervision during optimization while directly maximizing end-to-end spotting quality through dual complementary rewards. Coupled with \textbf{DEA}'s norm-decoupled directional prototype alignment and \textbf{PED}'s cross-attention-driven dense geometry recovery, SPaTS bridges the gap between the semantic reasoning capacity of large multimodal models and the fine-grained geometric rigor required for high-fidelity text spotting, offering a novel visual grounding paradigm for precision-demanding OCR applications in complex scenarios.

\FloatBarrier
\clearpage

\begin{acks}
This research is supported in part by the National Natural Science Foundation of China (Grant No. 62476093), the Natural Science Foundation of Guangdong Province (Grant No. 2026A1515012038), the China Postdoctoral Science Foundation (Grant No. 2026M791625), and the Postdoctoral Fellowship Program (Grade B) of the China Postdoctoral Science Foundation (Grant No. GZB20260386).
\end{acks}

\balance
\bibliographystyle{ACM-Reference-Format}
\bibliography{ready}


\begin{thebibliography}{58}


\ifx \showCODEN    \undefined \def \showCODEN     #1{\unskip}     \fi
\ifx \showISBNx    \undefined \def \showISBNx     #1{\unskip}     \fi
\ifx \showISBNxiii \undefined \def \showISBNxiii  #1{\unskip}     \fi
\ifx \showISSN     \undefined \def \showISSN      #1{\unskip}     \fi
\ifx \showLCCN     \undefined \def \showLCCN      #1{\unskip}     \fi
\ifx \shownote     \undefined \def \shownote      #1{#1}          \fi
\ifx \showarticletitle \undefined \def \showarticletitle #1{#1}   \fi
\ifx \showURL      \undefined \def \showURL       {\relax}        \fi
\providecommand\bibfield[2]{#2}
\providecommand\bibinfo[2]{#2}
\providecommand\natexlab[1]{#1}
\providecommand\showeprint[2][]{arXiv:#2}

\bibitem[{Anthropic}(2026)]%
        {anthropic2026claudesonnet46}
\bibfield{author}{\bibinfo{person}{{Anthropic}}.} \bibinfo{year}{2026}\natexlab{}.
\newblock \bibinfo{title}{Introducing Claude Sonnet 4.6}.
\newblock
\urldef\tempurl%
\url{https://www.anthropic.com/news/claude-sonnet-4-6}
\showURL{%
\tempurl}


\bibitem[Baek et~al\mbox{.}(2020)]%
        {baek2020character}
\bibfield{author}{\bibinfo{person}{Youngmin Baek}, \bibinfo{person}{Seung Shin}, \bibinfo{person}{Jeonghun Baek}, \bibinfo{person}{Sungrae Park}, \bibinfo{person}{Junyeop Lee}, \bibinfo{person}{Daehyun Nam}, {and} \bibinfo{person}{Hwalsuk Lee}.} \bibinfo{year}{2020}\natexlab{}.
\newblock \showarticletitle{Character Region Attention for Text Spotting}. In \bibinfo{booktitle}{\emph{Computer Vision - {ECCV} 2020 - 16th European Conference, Glasgow, UK, August 23-28, 2020, Proceedings, Part {XXIX}}} \emph{(\bibinfo{series}{Lecture Notes in Computer Science})}, \bibfield{editor}{\bibinfo{person}{Andrea Vedaldi}, \bibinfo{person}{Horst Bischof}, \bibinfo{person}{Thomas Brox}, {and} \bibinfo{person}{Jan{-}Michael Frahm}} (Eds.). \bibinfo{publisher}{Springer}, \bibinfo{pages}{504--521}.
\newblock
\href{https://doi.org/10.1007/978-3-030-58526-6\_30}{doi:\nolinkurl{10.1007/978-3-030-58526-6\_30}}


\bibitem[{Bytedance}(2026)]%
        {bytedance2026seed2.0}
\bibfield{author}{\bibinfo{person}{{Bytedance}}.} \bibinfo{year}{2026}\natexlab{}.
\newblock \bibinfo{title}{Seed 2.0 Official Launch}.
\newblock
\urldef\tempurl%
\url{https://seed.bytedance.com/zh/blog/seed-2-0-official-launch?view_from=content_recommend}
\showURL{%
\tempurl}


\bibitem[Chen et~al\mbox{.}(2023)]%
        {chen2023shikra}
\bibfield{author}{\bibinfo{person}{Keqin Chen}, \bibinfo{person}{Zhao Zhang}, \bibinfo{person}{Weili Zeng}, \bibinfo{person}{Richong Zhang}, \bibinfo{person}{Feng Zhu}, {and} \bibinfo{person}{Rui Zhao}.} \bibinfo{year}{2023}\natexlab{}.
\newblock \showarticletitle{Shikra: Unleashing Multimodal LLM's Referential Dialogue Magic}.
\newblock \bibinfo{journal}{\emph{CoRR}}  \bibinfo{volume}{abs/2306.15195} (\bibinfo{year}{2023}).
\newblock
\showeprint[arXiv]{2306.15195}
\href{https://doi.org/10.48550/ARXIV.2306.15195}{doi:\nolinkurl{10.48550/ARXIV.2306.15195}}


\bibitem[Chng and Chan(2017)]%
        {chng2017total}
\bibfield{author}{\bibinfo{person}{Chee~Kheng Chng} {and} \bibinfo{person}{Chee~Seng Chan}.} \bibinfo{year}{2017}\natexlab{}.
\newblock \showarticletitle{Total-Text: {A} Comprehensive Dataset for Scene Text Detection and Recognition}. In \bibinfo{booktitle}{\emph{14th {IAPR} International Conference on Document Analysis and Recognition, {ICDAR} 2017, Kyoto, Japan, November 9-15, 2017}}. \bibinfo{publisher}{{IEEE}}, \bibinfo{pages}{935--942}.
\newblock
\href{https://doi.org/10.1109/ICDAR.2017.157}{doi:\nolinkurl{10.1109/ICDAR.2017.157}}


\bibitem[Cui et~al\mbox{.}(2026)]%
        {cui2026paddleocr}
\bibfield{author}{\bibinfo{person}{Cheng Cui}, \bibinfo{person}{Ting Sun}, \bibinfo{person}{Suyin Liang}, \bibinfo{person}{Tingquan Gao}, \bibinfo{person}{Zelun Zhang}, \bibinfo{person}{Jiaxuan Liu}, \bibinfo{person}{Xueqing Wang}, \bibinfo{person}{Changda Zhou}, \bibinfo{person}{Hongen Liu}, \bibinfo{person}{Manhui Lin}, \bibinfo{person}{Yue Zhang}, \bibinfo{person}{Yubo Zhang}, \bibinfo{person}{Yi Liu}, \bibinfo{person}{Dianhai Yu}, {and} \bibinfo{person}{Yanjun Ma}.} \bibinfo{year}{2026}\natexlab{}.
\newblock \showarticletitle{PaddleOCR-VL-1.5: Towards a Multi-Task 0.9B {VLM} for Robust In-the-Wild Document Parsing}.
\newblock \bibinfo{journal}{\emph{CoRR}}  \bibinfo{volume}{abs/2601.21957} (\bibinfo{year}{2026}).
\newblock
\showeprint[arXiv]{2601.21957}
\href{https://doi.org/10.48550/ARXIV.2601.21957}{doi:\nolinkurl{10.48550/ARXIV.2601.21957}}


\bibitem[Cui et~al\mbox{.}(2025)]%
        {cui2025paddleocrvl}
\bibfield{author}{\bibinfo{person}{Cheng Cui}, \bibinfo{person}{Ting Sun}, \bibinfo{person}{Suyin Liang}, \bibinfo{person}{Tingquan Gao}, \bibinfo{person}{Zelun Zhang}, \bibinfo{person}{Jiaxuan Liu}, \bibinfo{person}{Xueqing Wang}, \bibinfo{person}{Changda Zhou}, \bibinfo{person}{Hongen Liu}, \bibinfo{person}{Manhui Lin}, \bibinfo{person}{Yue Zhang}, \bibinfo{person}{Yubo Zhang}, \bibinfo{person}{Handong Zheng}, \bibinfo{person}{Jing Zhang}, \bibinfo{person}{Jun Zhang}, \bibinfo{person}{Yi Liu}, \bibinfo{person}{Dianhai Yu}, {and} \bibinfo{person}{Yanjun Ma}.} \bibinfo{year}{2025}\natexlab{}.
\newblock \showarticletitle{PaddleOCR-VL: Boosting Multilingual Document Parsing via a 0.9B Ultra-Compact Vision-Language Model}.
\newblock \bibinfo{journal}{\emph{CoRR}}  \bibinfo{volume}{abs/2510.14528} (\bibinfo{year}{2025}).
\newblock
\showeprint[arXiv]{2510.14528}
\href{https://doi.org/10.48550/ARXIV.2510.14528}{doi:\nolinkurl{10.48550/ARXIV.2510.14528}}


\bibitem[Fang et~al\mbox{.}(2024)]%
        {Fang2024PUMA}
\bibfield{author}{\bibinfo{person}{Rongyao Fang}, \bibinfo{person}{Chengqi Duan}, \bibinfo{person}{Kun Wang}, \bibinfo{person}{Hao Li}, \bibinfo{person}{Hao Tian}, \bibinfo{person}{Xingyu Zeng}, \bibinfo{person}{Rui Zhao}, \bibinfo{person}{Jifeng Dai}, \bibinfo{person}{Hongsheng Li}, {and} \bibinfo{person}{Xihui Liu}.} \bibinfo{year}{2024}\natexlab{}.
\newblock \showarticletitle{{PUMA:} Empowering Unified {MLLM} with Multi-granular Visual Generation}.
\newblock \bibinfo{journal}{\emph{CoRR}}  \bibinfo{volume}{abs/2410.13861} (\bibinfo{year}{2024}).
\newblock
\showeprint[arXiv]{2410.13861}
\href{https://doi.org/10.48550/ARXIV.2410.13861}{doi:\nolinkurl{10.48550/ARXIV.2410.13861}}


\bibitem[Feng et~al\mbox{.}(2019)]%
        {feng2019textdragon}
\bibfield{author}{\bibinfo{person}{Wei Feng}, \bibinfo{person}{Wenhao He}, \bibinfo{person}{Fei Yin}, \bibinfo{person}{Xu{-}Yao Zhang}, {and} \bibinfo{person}{Cheng{-}Lin Liu}.} \bibinfo{year}{2019}\natexlab{}.
\newblock \showarticletitle{TextDragon: An End-to-End Framework for Arbitrary Shaped Text Spotting}. In \bibinfo{booktitle}{\emph{2019 {IEEE/CVF} International Conference on Computer Vision, {ICCV} 2019, Seoul, Korea (South), October 27 - November 2, 2019}}. \bibinfo{publisher}{{IEEE}}, \bibinfo{pages}{9075--9084}.
\newblock
\href{https://doi.org/10.1109/ICCV.2019.00917}{doi:\nolinkurl{10.1109/ICCV.2019.00917}}


\bibitem[{Gemini Team}(2025)]%
        {googleteam2025gemini3flash}
\bibfield{author}{\bibinfo{person}{{Gemini Team}}.} \bibinfo{year}{2025}\natexlab{}.
\newblock \bibinfo{title}{Gemini 3 Flash: frontier intelligence built for speed}.
\newblock
\urldef\tempurl%
\url{https://blog.google/products/gemini/gemini-3-flash/}
\showURL{%
\tempurl}


\bibitem[Guo et~al\mbox{.}(2025)]%
        {guo2025deepseek}
\bibfield{author}{\bibinfo{person}{Daya Guo}, \bibinfo{person}{Dejian Yang}, \bibinfo{person}{Haowei Zhang}, {et~al\mbox{.}}} \bibinfo{year}{2025}\natexlab{}.
\newblock \showarticletitle{{DeepSeek-R1} incentivizes reasoning in {LLMs} through reinforcement learning}.
\newblock \bibinfo{journal}{\emph{Nature}}  \bibinfo{volume}{645} (\bibinfo{year}{2025}), \bibinfo{pages}{633--638}.
\newblock
\href{https://doi.org/10.1038/s41586-025-09422-z}{doi:\nolinkurl{10.1038/s41586-025-09422-z}}


\bibitem[Huang et~al\mbox{.}(2025b)]%
        {jia2025visual}
\bibfield{author}{\bibinfo{person}{Jiaxing Huang}, \bibinfo{person}{Jingyi Zhang}, \bibinfo{person}{Kai Jiang}, \bibinfo{person}{Han Qiu}, \bibinfo{person}{Xiaoqin Zhang}, \bibinfo{person}{Ling Shao}, \bibinfo{person}{Shijian Lu}, {and} \bibinfo{person}{Dacheng Tao}.} \bibinfo{year}{2025}\natexlab{b}.
\newblock \showarticletitle{Visual Instruction Tuning towards General-Purpose Multimodal Large Language Model: {A} Survey}.
\newblock \bibinfo{journal}{\emph{Int. J. Comput. Vis.}} \bibinfo{volume}{133}, \bibinfo{number}{11} (\bibinfo{year}{2025}), \bibinfo{pages}{8151--8189}.
\newblock
\href{https://doi.org/10.1007/S11263-025-02572-7}{doi:\nolinkurl{10.1007/S11263-025-02572-7}}


\bibitem[Huang et~al\mbox{.}(2022)]%
        {huang2022swintextspotter}
\bibfield{author}{\bibinfo{person}{Mingxin Huang}, \bibinfo{person}{Yuliang Liu}, \bibinfo{person}{Zhenghao Peng}, \bibinfo{person}{Chongyu Liu}, \bibinfo{person}{Dahua Lin}, \bibinfo{person}{Shenggao Zhu}, \bibinfo{person}{Nicholas~Jing Yuan}, \bibinfo{person}{Kai Ding}, {and} \bibinfo{person}{Lianwen Jin}.} \bibinfo{year}{2022}\natexlab{}.
\newblock \showarticletitle{SwinTextSpotter: Scene Text Spotting via Better Synergy between Text Detection and Text Recognition}. In \bibinfo{booktitle}{\emph{{IEEE/CVF} Conference on Computer Vision and Pattern Recognition, {CVPR} 2022, New Orleans, LA, USA, June 18-24, 2022}}. \bibinfo{publisher}{{IEEE}}, \bibinfo{pages}{4583--4593}.
\newblock
\href{https://doi.org/10.1109/CVPR52688.2022.00455}{doi:\nolinkurl{10.1109/CVPR52688.2022.00455}}


\bibitem[Huang et~al\mbox{.}(2025a)]%
        {huang2024vlm_rl}
\bibfield{author}{\bibinfo{person}{Zilin Huang}, \bibinfo{person}{Zihao Sheng}, \bibinfo{person}{Yansong Qu}, \bibinfo{person}{Junwei You}, {and} \bibinfo{person}{Sikai Chen}.} \bibinfo{year}{2025}\natexlab{a}.
\newblock \showarticletitle{{VLM-RL:} {A} Unified Vision Language Models and Reinforcement Learning Framework for Safe Autonomous Driving}.
\newblock \bibinfo{journal}{\emph{Transportation Research Part C: Emerging Technologies}} (\bibinfo{year}{2025}).
\newblock
\href{https://doi.org/10.1016/j.trc.2025.105321}{doi:\nolinkurl{10.1016/j.trc.2025.105321}}


\bibitem[{Hugging Face}(2026)]%
        {hf2026trl}
\bibfield{author}{\bibinfo{person}{{Hugging Face}}.} \bibinfo{year}{2026}\natexlab{}.
\newblock \bibinfo{booktitle}{\emph{TRL: Transformers Reinforcement Learning}}.
\newblock
\shownote{Accessed: 2026-04-09}.
\newblock
\urldef\tempurl%
\url{https://huggingface.co/docs/trl}
\showURL{%
\tempurl}


\bibitem[Karatzas et~al\mbox{.}(2015)]%
        {karatzas2015icdar}
\bibfield{author}{\bibinfo{person}{Dimosthenis Karatzas}, \bibinfo{person}{Lluis Gomez{-}Bigorda}, \bibinfo{person}{Anguelos Nicolaou}, \bibinfo{person}{Suman~K. Ghosh}, \bibinfo{person}{Andrew~D. Bagdanov}, \bibinfo{person}{Masakazu Iwamura}, \bibinfo{person}{Jiri Matas}, \bibinfo{person}{Luk{\'{a}}s Neumann}, \bibinfo{person}{Vijay~Ramaseshan Chandrasekhar}, \bibinfo{person}{Shijian Lu}, \bibinfo{person}{Faisal Shafait}, \bibinfo{person}{Seiichi Uchida}, {and} \bibinfo{person}{Ernest Valveny}.} \bibinfo{year}{2015}\natexlab{}.
\newblock \showarticletitle{{ICDAR} 2015 competition on Robust Reading}. In \bibinfo{booktitle}{\emph{13th International Conference on Document Analysis and Recognition, {ICDAR} 2015, Nancy, France, August 23-26, 2015}}. \bibinfo{publisher}{{IEEE} Computer Society}, \bibinfo{pages}{1156--1160}.
\newblock
\href{https://doi.org/10.1109/ICDAR.2015.7333942}{doi:\nolinkurl{10.1109/ICDAR.2015.7333942}}


\bibitem[Karatzas et~al\mbox{.}(2013)]%
        {karatzas2013icdar}
\bibfield{author}{\bibinfo{person}{Dimosthenis Karatzas}, \bibinfo{person}{Faisal Shafait}, \bibinfo{person}{Seiichi Uchida}, \bibinfo{person}{Masakazu Iwamura}, \bibinfo{person}{Lluis~Gomez i Bigorda}, \bibinfo{person}{Sergi~Robles Mestre}, \bibinfo{person}{Joan Mas}, \bibinfo{person}{David~Fern{\'{a}}ndez Mota}, \bibinfo{person}{Jon Almaz{\'{a}}n}, {and} \bibinfo{person}{Llu{\'{\i}}s{-}Pere de~las Heras}.} \bibinfo{year}{2013}\natexlab{}.
\newblock \showarticletitle{{ICDAR} 2013 Robust Reading Competition}. In \bibinfo{booktitle}{\emph{12th International Conference on Document Analysis and Recognition, {ICDAR} 2013, Washington, DC, USA, August 25-28, 2013}}. \bibinfo{publisher}{{IEEE} Computer Society}, \bibinfo{pages}{1484--1493}.
\newblock
\href{https://doi.org/10.1109/ICDAR.2013.221}{doi:\nolinkurl{10.1109/ICDAR.2013.221}}


\bibitem[Lei et~al\mbox{.}(2016)]%
        {lei2016rationalizing}
\bibfield{author}{\bibinfo{person}{Tao Lei}, \bibinfo{person}{Regina Barzilay}, {and} \bibinfo{person}{Tommi~S. Jaakkola}.} \bibinfo{year}{2016}\natexlab{}.
\newblock \showarticletitle{Rationalizing Neural Predictions}. In \bibinfo{booktitle}{\emph{Proceedings of the 2016 Conference on Empirical Methods in Natural Language Processing, {EMNLP} 2016, Austin, Texas, USA, November 1-4, 2016}}, \bibfield{editor}{\bibinfo{person}{Jian Su}, \bibinfo{person}{Xavier Carreras}, {and} \bibinfo{person}{Kevin Duh}} (Eds.). \bibinfo{publisher}{The Association for Computational Linguistics}, \bibinfo{pages}{107--117}.
\newblock
\href{https://doi.org/10.18653/V1/D16-1011}{doi:\nolinkurl{10.18653/V1/D16-1011}}


\bibitem[Li et~al\mbox{.}(2024)]%
        {li2024monkey}
\bibfield{author}{\bibinfo{person}{Zhang Li}, \bibinfo{person}{Biao Yang}, \bibinfo{person}{Qiang Liu}, \bibinfo{person}{Zhiyin Ma}, \bibinfo{person}{Shuo Zhang}, \bibinfo{person}{Jingxu Yang}, \bibinfo{person}{Yabo Sun}, \bibinfo{person}{Yuliang Liu}, {and} \bibinfo{person}{Xiang Bai}.} \bibinfo{year}{2024}\natexlab{}.
\newblock \showarticletitle{Monkey: Image Resolution and Text Label are Important Things for Large Multi-Modal Models}. In \bibinfo{booktitle}{\emph{{IEEE/CVF} Conference on Computer Vision and Pattern Recognition, {CVPR} 2024, Seattle, WA, USA, June 16-22, 2024}}. \bibinfo{publisher}{{IEEE}}, \bibinfo{pages}{26753--26763}.
\newblock
\href{https://doi.org/10.1109/CVPR52733.2024.02527}{doi:\nolinkurl{10.1109/CVPR52733.2024.02527}}


\bibitem[Liao et~al\mbox{.}(2021)]%
        {lyu2018mask}
\bibfield{author}{\bibinfo{person}{Minghui Liao}, \bibinfo{person}{Pengyuan Lyu}, \bibinfo{person}{Minghang He}, \bibinfo{person}{Cong Yao}, \bibinfo{person}{Wenhao Wu}, {and} \bibinfo{person}{Xiang Bai}.} \bibinfo{year}{2021}\natexlab{}.
\newblock \showarticletitle{Mask TextSpotter: An End-to-End Trainable Neural Network for Spotting Text with Arbitrary Shapes}.
\newblock \bibinfo{journal}{\emph{{IEEE} Trans. Pattern Anal. Mach. Intell.}} \bibinfo{volume}{43}, \bibinfo{number}{2} (\bibinfo{year}{2021}), \bibinfo{pages}{532--548}.
\newblock
\href{https://doi.org/10.1109/TPAMI.2019.2937086}{doi:\nolinkurl{10.1109/TPAMI.2019.2937086}}


\bibitem[Liao et~al\mbox{.}(2023)]%
        {liao2020real}
\bibfield{author}{\bibinfo{person}{Minghui Liao}, \bibinfo{person}{Zhisheng Zou}, \bibinfo{person}{Zhaoyi Wan}, \bibinfo{person}{Cong Yao}, {and} \bibinfo{person}{Xiang Bai}.} \bibinfo{year}{2023}\natexlab{}.
\newblock \showarticletitle{Real-Time Scene Text Detection With Differentiable Binarization and Adaptive Scale Fusion}.
\newblock \bibinfo{journal}{\emph{{IEEE} Trans. Pattern Anal. Mach. Intell.}} \bibinfo{volume}{45}, \bibinfo{number}{1} (\bibinfo{year}{2023}), \bibinfo{pages}{919--931}.
\newblock
\href{https://doi.org/10.1109/TPAMI.2022.3155612}{doi:\nolinkurl{10.1109/TPAMI.2022.3155612}}


\bibitem[Liu et~al\mbox{.}(2018)]%
        {liu2018fots}
\bibfield{author}{\bibinfo{person}{Xuebo Liu}, \bibinfo{person}{Ding Liang}, \bibinfo{person}{Shi Yan}, \bibinfo{person}{Dagui Chen}, \bibinfo{person}{Yu Qiao}, {and} \bibinfo{person}{Junjie Yan}.} \bibinfo{year}{2018}\natexlab{}.
\newblock \showarticletitle{{FOTS:} Fast Oriented Text Spotting With a Unified Network}. In \bibinfo{booktitle}{\emph{2018 {IEEE} Conference on Computer Vision and Pattern Recognition, {CVPR} 2018, Salt Lake City, UT, USA, June 18-22, 2018}}. \bibinfo{publisher}{Computer Vision Foundation / {IEEE} Computer Society}, \bibinfo{pages}{5676--5685}.
\newblock
\href{https://doi.org/10.1109/CVPR.2018.00595}{doi:\nolinkurl{10.1109/CVPR.2018.00595}}


\bibitem[Liu et~al\mbox{.}(2020)]%
        {liu2020abcnet}
\bibfield{author}{\bibinfo{person}{Yuliang Liu}, \bibinfo{person}{Hao Chen}, \bibinfo{person}{Chunhua Shen}, \bibinfo{person}{Tong He}, \bibinfo{person}{Lianwen Jin}, {and} \bibinfo{person}{Liangwei Wang}.} \bibinfo{year}{2020}\natexlab{}.
\newblock \showarticletitle{ABCNet: Real-Time Scene Text Spotting With Adaptive Bezier-Curve Network}. In \bibinfo{booktitle}{\emph{2020 {IEEE/CVF} Conference on Computer Vision and Pattern Recognition, {CVPR} 2020, Seattle, WA, USA, June 13-19, 2020}}. \bibinfo{publisher}{Computer Vision Foundation / {IEEE}}, \bibinfo{pages}{9806--9815}.
\newblock
\href{https://doi.org/10.1109/CVPR42600.2020.00983}{doi:\nolinkurl{10.1109/CVPR42600.2020.00983}}


\bibitem[Liu et~al\mbox{.}(2019)]%
        {liu2019curved}
\bibfield{author}{\bibinfo{person}{Yuliang Liu}, \bibinfo{person}{Lianwen Jin}, \bibinfo{person}{Shuaitao Zhang}, \bibinfo{person}{Canjie Luo}, {and} \bibinfo{person}{Sheng Zhang}.} \bibinfo{year}{2019}\natexlab{}.
\newblock \showarticletitle{Curved scene text detection via transverse and longitudinal sequence connection}.
\newblock \bibinfo{journal}{\emph{Pattern Recognit.}}  \bibinfo{volume}{90} (\bibinfo{year}{2019}), \bibinfo{pages}{337--345}.
\newblock
\href{https://doi.org/10.1016/J.PATCOG.2019.02.002}{doi:\nolinkurl{10.1016/J.PATCOG.2019.02.002}}


\bibitem[Liu et~al\mbox{.}(2022)]%
        {liu2022abcnetv2}
\bibfield{author}{\bibinfo{person}{Yuliang Liu}, \bibinfo{person}{Chunhua Shen}, \bibinfo{person}{Lianwen Jin}, \bibinfo{person}{Tong He}, \bibinfo{person}{Peng Chen}, \bibinfo{person}{Chongyu Liu}, {and} \bibinfo{person}{Hao Chen}.} \bibinfo{year}{2022}\natexlab{}.
\newblock \showarticletitle{ABCNet v2: Adaptive Bezier-Curve Network for Real-Time End-to-End Text Spotting}.
\newblock \bibinfo{journal}{\emph{{IEEE} Trans. Pattern Anal. Mach. Intell.}} \bibinfo{volume}{44}, \bibinfo{number}{11} (\bibinfo{year}{2022}), \bibinfo{pages}{8048--8064}.
\newblock
\href{https://doi.org/10.1109/TPAMI.2021.3107437}{doi:\nolinkurl{10.1109/TPAMI.2021.3107437}}


\bibitem[Liu et~al\mbox{.}(2023)]%
        {liu2023sptsv2}
\bibfield{author}{\bibinfo{person}{Yuliang Liu}, \bibinfo{person}{Jiaxin Zhang}, \bibinfo{person}{Dezhi Peng}, \bibinfo{person}{Mingxin Huang}, \bibinfo{person}{Xinyu Wang}, \bibinfo{person}{Jingqun Tang}, \bibinfo{person}{Can Huang}, \bibinfo{person}{Dahua Lin}, \bibinfo{person}{Chunhua Shen}, \bibinfo{person}{Xiang Bai}, {and} \bibinfo{person}{Lianwen Jin}.} \bibinfo{year}{2023}\natexlab{}.
\newblock \showarticletitle{{SPTS} v2: Single-Point Scene Text Spotting}.
\newblock \bibinfo{journal}{\emph{{IEEE} Trans. Pattern Anal. Mach. Intell.}} \bibinfo{volume}{45}, \bibinfo{number}{12} (\bibinfo{year}{2023}), \bibinfo{pages}{15665--15679}.
\newblock
\href{https://doi.org/10.1109/TPAMI.2023.3312285}{doi:\nolinkurl{10.1109/TPAMI.2023.3312285}}


\bibitem[Liu et~al\mbox{.}(2025)]%
        {liu2025visual_rft}
\bibfield{author}{\bibinfo{person}{Ziyu Liu}, \bibinfo{person}{Zeyi Sun}, \bibinfo{person}{Yuhang Zang}, \bibinfo{person}{Xiaoyi Dong}, \bibinfo{person}{Yuhang Cao}, \bibinfo{person}{Haodong Duan}, \bibinfo{person}{Dahua Lin}, {and} \bibinfo{person}{Jiaqi Wang}.} \bibinfo{year}{2025}\natexlab{}.
\newblock \showarticletitle{Visual-RFT: Visual Reinforcement Fine-Tuning}. In \bibinfo{booktitle}{\emph{Proceedings of the IEEE/CVF International Conference on Computer Vision (ICCV)}}. \bibinfo{pages}{2034--2044}.
\newblock
\urldef\tempurl%
\url{https://openaccess.thecvf.com/content/ICCV2025/html/Liu_Visual-RFT_Visual_Reinforcement_Fine-Tuning_ICCV_2025_paper.html}
\showURL{%
\tempurl}


\bibitem[Long et~al\mbox{.}(2021)]%
        {long2021scene}
\bibfield{author}{\bibinfo{person}{Shangbang Long}, \bibinfo{person}{Xin He}, {and} \bibinfo{person}{Cong Yao}.} \bibinfo{year}{2021}\natexlab{}.
\newblock \showarticletitle{Scene Text Detection and Recognition: The Deep Learning Era}.
\newblock \bibinfo{journal}{\emph{Int. J. Comput. Vis.}} \bibinfo{volume}{129}, \bibinfo{number}{1} (\bibinfo{year}{2021}), \bibinfo{pages}{161--184}.
\newblock
\href{https://doi.org/10.1007/S11263-020-01369-0}{doi:\nolinkurl{10.1007/S11263-020-01369-0}}


\bibitem[Ma et~al\mbox{.}(2024)]%
        {ma2024groma}
\bibfield{author}{\bibinfo{person}{Chuofan Ma}, \bibinfo{person}{Yi Jiang}, \bibinfo{person}{Jiannan Wu}, \bibinfo{person}{Zehuan Yuan}, {and} \bibinfo{person}{Xiaojuan Qi}.} \bibinfo{year}{2024}\natexlab{}.
\newblock \showarticletitle{Groma: Localized Visual Tokenization for Grounding Multimodal Large Language Models}. In \bibinfo{booktitle}{\emph{Computer Vision - {ECCV} 2024 - 18th European Conference, Milan, Italy, September 29-October 4, 2024, Proceedings, Part {VI}}} \emph{(\bibinfo{series}{Lecture Notes in Computer Science})}, \bibfield{editor}{\bibinfo{person}{Ales Leonardis}, \bibinfo{person}{Elisa Ricci}, \bibinfo{person}{Stefan Roth}, \bibinfo{person}{Olga Russakovsky}, \bibinfo{person}{Torsten Sattler}, {and} \bibinfo{person}{G{\"{u}}l Varol}} (Eds.). \bibinfo{publisher}{Springer}, \bibinfo{pages}{417--435}.
\newblock
\href{https://doi.org/10.1007/978-3-031-72658-3\_24}{doi:\nolinkurl{10.1007/978-3-031-72658-3\_24}}


\bibitem[Ma et~al\mbox{.}(2025)]%
        {ma2025clawmachine}
\bibfield{author}{\bibinfo{person}{Tianren Ma}, \bibinfo{person}{Lingxi Xie}, \bibinfo{person}{Yunjie Tian}, \bibinfo{person}{Boyu Yang}, {and} \bibinfo{person}{Qixiang Ye}.} \bibinfo{year}{2025}\natexlab{}.
\newblock \showarticletitle{ClawMachine: Learning to Fetch Visual Tokens for Referential Comprehension}. In \bibinfo{booktitle}{\emph{The Thirteenth International Conference on Learning Representations, {ICLR} 2025, Singapore, April 24-28, 2025}}. \bibinfo{publisher}{OpenReview.net}.
\newblock
\urldef\tempurl%
\url{https://openreview.net/forum?id=TOtk9dTYGG}
\showURL{%
\tempurl}


\bibitem[Ma et~al\mbox{.}(2026)]%
        {Ma2026One2SeqOW}
\bibfield{author}{\bibinfo{person}{Zhibin Ma}, \bibinfo{person}{Pengwen Dai}, \bibinfo{person}{Wei Zhuo}, {and} \bibinfo{person}{Xugong Qin}.} \bibinfo{year}{2026}\natexlab{}.
\newblock \showarticletitle{One2Seq: One-Token Wise Decoder for Efficient Scene Text Recognition}.
\newblock \bibinfo{journal}{\emph{Proceedings of the AAAI Conference on Artificial Intelligence}} (\bibinfo{year}{2026}).
\newblock
\urldef\tempurl%
\url{https://api.semanticscholar.org/CorpusID:286656590}
\showURL{%
\tempurl}


\bibitem[Nayef et~al\mbox{.}(2017)]%
        {nayef2017icdar}
\bibfield{author}{\bibinfo{person}{Nibal Nayef}, \bibinfo{person}{Fei Yin}, \bibinfo{person}{Imen Bizid}, \bibinfo{person}{Hyunsoo Choi}, \bibinfo{person}{Yuan Feng}, \bibinfo{person}{Dimosthenis Karatzas}, \bibinfo{person}{Zhenbo Luo}, \bibinfo{person}{Umapada Pal}, \bibinfo{person}{Christophe Rigaud}, \bibinfo{person}{Joseph Chazalon}, \bibinfo{person}{Wafa Khlif}, \bibinfo{person}{Muhammad~Muzzamil Luqman}, \bibinfo{person}{Jean{-}Christophe Burie}, \bibinfo{person}{Cheng{-}Lin Liu}, {and} \bibinfo{person}{Jean{-}Marc Ogier}.} \bibinfo{year}{2017}\natexlab{}.
\newblock \showarticletitle{{ICDAR2017} Robust Reading Challenge on Multi-Lingual Scene Text Detection and Script Identification - {RRC-MLT}}. In \bibinfo{booktitle}{\emph{14th {IAPR} International Conference on Document Analysis and Recognition, {ICDAR} 2017, Kyoto, Japan, November 9-15, 2017}}. \bibinfo{publisher}{{IEEE}}, \bibinfo{pages}{1454--1459}.
\newblock
\href{https://doi.org/10.1109/ICDAR.2017.237}{doi:\nolinkurl{10.1109/ICDAR.2017.237}}


\bibitem[{OpenAI}(2026)]%
        {openai2026gpt54}
\bibfield{author}{\bibinfo{person}{{OpenAI}}.} \bibinfo{year}{2026}\natexlab{}.
\newblock \bibinfo{title}{Introducing GPT-5.4}.
\newblock
\urldef\tempurl%
\url{https://openai.com/index/introducing-gpt-5-4/}
\showURL{%
\tempurl}


\bibitem[Ouyang et~al\mbox{.}(2022)]%
        {ouyang2022training}
\bibfield{author}{\bibinfo{person}{Long Ouyang}, \bibinfo{person}{Jeffrey Wu}, \bibinfo{person}{Xu Jiang}, \bibinfo{person}{Diogo Almeida}, \bibinfo{person}{Carroll~L. Wainwright}, \bibinfo{person}{Pamela Mishkin}, \bibinfo{person}{Chong Zhang}, \bibinfo{person}{Sandhini Agarwal}, \bibinfo{person}{Katarina Slama}, \bibinfo{person}{Alex Ray}, \bibinfo{person}{John Schulman}, \bibinfo{person}{Jacob Hilton}, \bibinfo{person}{Fraser Kelton}, \bibinfo{person}{Luke Miller}, \bibinfo{person}{Maddie Simens}, \bibinfo{person}{Amanda Askell}, \bibinfo{person}{Peter Welinder}, \bibinfo{person}{Paul~F. Christiano}, \bibinfo{person}{Jan Leike}, {and} \bibinfo{person}{Ryan Lowe}.} \bibinfo{year}{2022}\natexlab{}.
\newblock \showarticletitle{Training language models to follow instructions with human feedback}. In \bibinfo{booktitle}{\emph{Advances in Neural Information Processing Systems 35: Annual Conference on Neural Information Processing Systems 2022, NeurIPS 2022, New Orleans, LA, USA, November 28 - December 9, 2022}}, \bibfield{editor}{\bibinfo{person}{Sanmi Koyejo}, \bibinfo{person}{S.~Mohamed}, \bibinfo{person}{A.~Agarwal}, \bibinfo{person}{Danielle Belgrave}, \bibinfo{person}{K.~Cho}, {and} \bibinfo{person}{A.~Oh}} (Eds.).
\newblock
\urldef\tempurl%
\url{http://papers.nips.cc/paper\_files/paper/2022/hash/b1efde53be364a73914f58805a001731-Abstract-Conference.html}
\showURL{%
\tempurl}


\bibitem[Peng et~al\mbox{.}(2022)]%
        {peng2022spts}
\bibfield{author}{\bibinfo{person}{Dezhi Peng}, \bibinfo{person}{Xinyu Wang}, \bibinfo{person}{Yuliang Liu}, \bibinfo{person}{Jiaxin Zhang}, \bibinfo{person}{Mingxin Huang}, \bibinfo{person}{Songxuan Lai}, \bibinfo{person}{Jing Li}, \bibinfo{person}{Shenggao Zhu}, \bibinfo{person}{Dahua Lin}, \bibinfo{person}{Chunhua Shen}, \bibinfo{person}{Xiang Bai}, {and} \bibinfo{person}{Lianwen Jin}.} \bibinfo{year}{2022}\natexlab{}.
\newblock \showarticletitle{{SPTS:} Single-Point Text Spotting}. In \bibinfo{booktitle}{\emph{{MM} '22: The 30th {ACM} International Conference on Multimedia, Lisboa, Portugal, October 10 - 14, 2022}}, \bibfield{editor}{\bibinfo{person}{Jo{\~{a}}o Magalh{\~{a}}es}, \bibinfo{person}{Alberto~Del Bimbo}, \bibinfo{person}{Shin'ichi Satoh}, \bibinfo{person}{Nicu Sebe}, \bibinfo{person}{Xavier Alameda{-}Pineda}, \bibinfo{person}{Qin Jin}, \bibinfo{person}{Vincent Oria}, {and} \bibinfo{person}{Laura Toni}} (Eds.). \bibinfo{publisher}{{ACM}}, \bibinfo{pages}{4272--4281}.
\newblock
\href{https://doi.org/10.1145/3503161.3547942}{doi:\nolinkurl{10.1145/3503161.3547942}}


\bibitem[Peng et~al\mbox{.}(2024)]%
        {peng2023kosmos}
\bibfield{author}{\bibinfo{person}{Zhiliang Peng}, \bibinfo{person}{Wenhui Wang}, \bibinfo{person}{Li Dong}, \bibinfo{person}{Yaru Hao}, \bibinfo{person}{Shaohan Huang}, \bibinfo{person}{Shuming Ma}, {and} \bibinfo{person}{Furu Wei}.} \bibinfo{year}{2024}\natexlab{}.
\newblock \showarticletitle{Kosmos-2: Grounding Multimodal Large Language Models to the World}. In \bibinfo{booktitle}{\emph{The Twelfth International Conference on Learning Representations, {ICLR} 2024, Vienna, Austria, May 7-11, 2024}}.
\newblock
\urldef\tempurl%
\url{https://openreview.net/forum?id=lLmqxkfSIw}
\showURL{%
\tempurl}


\bibitem[Ren et~al\mbox{.}(2017)]%
        {ren2017faster}
\bibfield{author}{\bibinfo{person}{Shaoqing Ren}, \bibinfo{person}{Kaiming He}, \bibinfo{person}{Ross~B. Girshick}, {and} \bibinfo{person}{Jian Sun}.} \bibinfo{year}{2017}\natexlab{}.
\newblock \showarticletitle{Faster {R-CNN:} Towards Real-Time Object Detection with Region Proposal Networks}.
\newblock \bibinfo{journal}{\emph{{IEEE} Trans. Pattern Anal. Mach. Intell.}} \bibinfo{volume}{39}, \bibinfo{number}{6} (\bibinfo{year}{2017}), \bibinfo{pages}{1137--1149}.
\newblock
\href{https://doi.org/10.1109/TPAMI.2016.2577031}{doi:\nolinkurl{10.1109/TPAMI.2016.2577031}}


\bibitem[Shao et~al\mbox{.}(2024)]%
        {shao2024deepseekmath}
\bibfield{author}{\bibinfo{person}{Zhihong Shao}, \bibinfo{person}{Peiyi Wang}, \bibinfo{person}{Qihao Zhu}, \bibinfo{person}{Runxin Xu}, \bibinfo{person}{Junxiao Song}, \bibinfo{person}{Mingchuan Zhang}, \bibinfo{person}{Y.~K. Li}, \bibinfo{person}{Y. Wu}, {and} \bibinfo{person}{Daya Guo}.} \bibinfo{year}{2024}\natexlab{}.
\newblock \showarticletitle{DeepSeekMath: Pushing the Limits of Mathematical Reasoning in Open Language Models}.
\newblock \bibinfo{journal}{\emph{CoRR}}  \bibinfo{volume}{abs/2402.03300} (\bibinfo{year}{2024}).
\newblock
\showeprint[arXiv]{2402.03300}
\href{https://doi.org/10.48550/ARXIV.2402.03300}{doi:\nolinkurl{10.48550/ARXIV.2402.03300}}


\bibitem[Shen et~al\mbox{.}(2025)]%
        {shen2025vlm_r1}
\bibfield{author}{\bibinfo{person}{Haozhan Shen}, \bibinfo{person}{Peng Liu}, \bibinfo{person}{Jingcheng Li}, \bibinfo{person}{Chunxin Fang}, \bibinfo{person}{Yibo Ma}, \bibinfo{person}{Jiajia Liao}, \bibinfo{person}{Qiaoli Shen}, \bibinfo{person}{Zilun Zhang}, \bibinfo{person}{Kangjia Zhao}, \bibinfo{person}{Qianqian Zhang}, \bibinfo{person}{Ruochen Xu}, {and} \bibinfo{person}{Tiancheng Zhao}.} \bibinfo{year}{2025}\natexlab{}.
\newblock \showarticletitle{{VLM-R1:} {A} Stable and Generalizable R1-style Large Vision-Language Model}.
\newblock \bibinfo{journal}{\emph{CoRR}}  \bibinfo{volume}{abs/2504.07615} (\bibinfo{year}{2025}).
\newblock
\showeprint[arXiv]{2504.07615}
\href{https://doi.org/10.48550/ARXIV.2504.07615}{doi:\nolinkurl{10.48550/ARXIV.2504.07615}}


\bibitem[Su et~al\mbox{.}(2026)]%
        {su2026patch}
\bibfield{author}{\bibinfo{person}{Yongyi Su}, \bibinfo{person}{Haojie Zhang}, \bibinfo{person}{Shijie Li}, \bibinfo{person}{Nanqing Liu}, \bibinfo{person}{Jingyi Liao}, \bibinfo{person}{Junyi Pan}, \bibinfo{person}{Yuan Liu}, \bibinfo{person}{Xiaofen Xing}, \bibinfo{person}{Chong Sun}, \bibinfo{person}{Chen Li}, \bibinfo{person}{Nancy~F. Chen}, \bibinfo{person}{Shuicheng Yan}, \bibinfo{person}{Xulei Yang}, {and} \bibinfo{person}{Xun Xu}.} \bibinfo{year}{2026}\natexlab{}.
\newblock \showarticletitle{Patch-as-Decodable-Token: Towards Unified Multi-Modal Vision Tasks in MLLMs}. In \bibinfo{booktitle}{\emph{The International Conference on Learning Representations (ICLR)}}.
\newblock
\urldef\tempurl%
\url{https://openreview.net/forum?id=xF0Dcmvsl0}
\showURL{%
\tempurl}


\bibitem[Team et~al\mbox{.}(2025)]%
        {hunyuanvisionteam2025hunyuanocr}
\bibfield{author}{\bibinfo{person}{Hunyuan~Vision Team}, \bibinfo{person}{Pengyuan Lyu}, \bibinfo{person}{Xingyu Wan}, \bibinfo{person}{Gengluo Li}, \bibinfo{person}{Shangpin Peng}, \bibinfo{person}{Weinong Wang}, \bibinfo{person}{Liang Wu}, \bibinfo{person}{Huawen Shen}, \bibinfo{person}{Yu Zhou}, \bibinfo{person}{Canhui Tang}, \bibinfo{person}{Qi Yang}, \bibinfo{person}{Qiming Peng}, \bibinfo{person}{Bin Luo}, \bibinfo{person}{Hower Yang}, \bibinfo{person}{Xinsong Zhang}, \bibinfo{person}{Jinnian Zhang}, \bibinfo{person}{Houwen Peng}, \bibinfo{person}{Hongming Yang}, \bibinfo{person}{Senhao Xie}, \bibinfo{person}{Longsha Zhou}, \bibinfo{person}{Ge Pei}, \bibinfo{person}{Binghong Wu}, \bibinfo{person}{Rui Yan}, \bibinfo{person}{Kan Wu}, \bibinfo{person}{Jieneng Yang}, \bibinfo{person}{Bochao Wang}, \bibinfo{person}{Kai Liu}, \bibinfo{person}{Jianchen Zhu}, \bibinfo{person}{Jie Jiang}, \bibinfo{person}{Linus}, \bibinfo{person}{Han Hu}, {and} \bibinfo{person}{Chengquan Zhang}.} \bibinfo{year}{2025}\natexlab{}.
\newblock \showarticletitle{HunyuanOCR Technical Report}.
\newblock \bibinfo{journal}{\emph{CoRR}}  \bibinfo{volume}{abs/2511.19575} (\bibinfo{year}{2025}).
\newblock
\showeprint[arXiv]{2511.19575}
\href{https://doi.org/10.48550/ARXIV.2511.19575}{doi:\nolinkurl{10.48550/ARXIV.2511.19575}}


\bibitem[Team(2026a)]%
        {moonshotai2026kimik25}
\bibfield{author}{\bibinfo{person}{Kimi Team}.} \bibinfo{year}{2026}\natexlab{a}.
\newblock \showarticletitle{Kimi {K2.5:} Visual Agentic Intelligence}.
\newblock \bibinfo{journal}{\emph{CoRR}}  \bibinfo{volume}{abs/2602.02276} (\bibinfo{year}{2026}).
\newblock
\showeprint[arXiv]{2602.02276}
\href{https://doi.org/10.48550/ARXIV.2602.02276}{doi:\nolinkurl{10.48550/ARXIV.2602.02276}}


\bibitem[Team(2025)]%
        {bai2025qwen3vl}
\bibfield{author}{\bibinfo{person}{Qwen Team}.} \bibinfo{year}{2025}\natexlab{}.
\newblock \showarticletitle{Qwen3-VL Technical Report}.
\newblock \bibinfo{journal}{\emph{CoRR}}  \bibinfo{volume}{abs/2511.21631} (\bibinfo{year}{2025}).
\newblock
\showeprint[arXiv]{2511.21631}
\href{https://doi.org/10.48550/ARXIV.2511.21631}{doi:\nolinkurl{10.48550/ARXIV.2511.21631}}


\bibitem[Team(2026b)]%
        {qwenteam2026qwen35}
\bibfield{author}{\bibinfo{person}{Qwen Team}.} \bibinfo{year}{2026}\natexlab{b}.
\newblock \showarticletitle{Qwen3. 5: Towards native multimodal agents}.
\newblock \bibinfo{journal}{\emph{URL: https://qwen. ai/blog}} (\bibinfo{year}{2026}).
\newblock


\bibitem[Tu et~al\mbox{.}(2024)]%
        {tu2024unicorns}
\bibfield{author}{\bibinfo{person}{Haoqin Tu}, \bibinfo{person}{Chenhang Cui}, \bibinfo{person}{Zijun Wang}, \bibinfo{person}{Yiyang Zhou}, \bibinfo{person}{Bingchen Zhao}, \bibinfo{person}{Junlin Han}, \bibinfo{person}{Wangchunshu Zhou}, \bibinfo{person}{Huaxiu Yao}, {and} \bibinfo{person}{Cihang Xie}.} \bibinfo{year}{2024}\natexlab{}.
\newblock \showarticletitle{How Many Unicorns Are in This Image? {A} Safety Evaluation Benchmark for Vision LLMs}. In \bibinfo{booktitle}{\emph{Proceedings of the European Conference on Computer Vision (ECCV)}}.
\newblock
\urldef\tempurl%
\url{https://eccv.ecva.net/virtual/2024/poster/1855}
\showURL{%
\tempurl}


\bibitem[Wan et~al\mbox{.}(2024)]%
        {wan2024omniparser}
\bibfield{author}{\bibinfo{person}{Jianqiang Wan}, \bibinfo{person}{Sibo Song}, \bibinfo{person}{Wenwen Yu}, \bibinfo{person}{Yuliang Liu}, \bibinfo{person}{Wenqing Cheng}, \bibinfo{person}{Fei Huang}, \bibinfo{person}{Xiang Bai}, \bibinfo{person}{Cong Yao}, {and} \bibinfo{person}{Zhibo Yang}.} \bibinfo{year}{2024}\natexlab{}.
\newblock \showarticletitle{{OMNIPARSER:} {A} Unified Framework for Text Spotting, Key Information Extraction and Table Recognition}. In \bibinfo{booktitle}{\emph{{IEEE/CVF} Conference on Computer Vision and Pattern Recognition, {CVPR} 2024, Seattle, WA, USA, June 16-22, 2024}}. \bibinfo{publisher}{{IEEE}}, \bibinfo{pages}{15641--15653}.
\newblock
\href{https://doi.org/10.1109/CVPR52733.2024.01481}{doi:\nolinkurl{10.1109/CVPR52733.2024.01481}}


\bibitem[Wang et~al\mbox{.}(2020)]%
        {wang2020all}
\bibfield{author}{\bibinfo{person}{Hao Wang}, \bibinfo{person}{Pu Lu}, \bibinfo{person}{Hui Zhang}, \bibinfo{person}{Mingkun Yang}, \bibinfo{person}{Xiang Bai}, \bibinfo{person}{Yongchao Xu}, \bibinfo{person}{Mengchao He}, \bibinfo{person}{Yongpan Wang}, {and} \bibinfo{person}{Wenyu Liu}.} \bibinfo{year}{2020}\natexlab{}.
\newblock \showarticletitle{All You Need Is Boundary: Toward Arbitrary-Shaped Text Spotting}. In \bibinfo{booktitle}{\emph{The Thirty-Fourth {AAAI} Conference on Artificial Intelligence, {AAAI} 2020, The Thirty-Second Innovative Applications of Artificial Intelligence Conference, {IAAI} 2020, The Tenth {AAAI} Symposium on Educational Advances in Artificial Intelligence, {EAAI} 2020, New York, NY, USA, February 7-12, 2020}}. \bibinfo{publisher}{{AAAI} Press}, \bibinfo{pages}{12160--12167}.
\newblock
\href{https://doi.org/10.1609/AAAI.V34I07.6896}{doi:\nolinkurl{10.1609/AAAI.V34I07.6896}}


\bibitem[Wang et~al\mbox{.}(2025)]%
        {wang2025internvl3_5}
\bibfield{author}{\bibinfo{person}{Weiyun Wang}, \bibinfo{person}{Zhangwei Gao}, \bibinfo{person}{Lixin Gu}, \bibinfo{person}{Hengjun Pu}, \bibinfo{person}{Long Cui}, \bibinfo{person}{Xingguang Wei}, \bibinfo{person}{Zhaoyang Liu}, \bibinfo{person}{Linglin Jing}, \bibinfo{person}{Shenglong Ye}, \bibinfo{person}{Jie Shao}, \bibinfo{person}{Zhaokai Wang}, \bibinfo{person}{Zhe Chen}, \bibinfo{person}{Hongjie Zhang}, \bibinfo{person}{Ganlin Yang}, \bibinfo{person}{Haomin Wang}, \bibinfo{person}{Qi Wei}, \bibinfo{person}{Jinhui Yin}, \bibinfo{person}{Wenhao Li}, \bibinfo{person}{Erfei Cui}, \bibinfo{person}{Guanzhou Chen}, \bibinfo{person}{Zichen Ding}, \bibinfo{person}{Changyao Tian}, \bibinfo{person}{Zhenyu Wu}, \bibinfo{person}{JingJing Xie}, \bibinfo{person}{Zehao Li}, \bibinfo{person}{Bowen Yang}, \bibinfo{person}{Yuchen Duan}, \bibinfo{person}{Xuehui Wang}, \bibinfo{person}{Zhi Hou}, \bibinfo{person}{Haoran Hao}, \bibinfo{person}{Tianyi Zhang}, \bibinfo{person}{Songze Li}, \bibinfo{person}{Xiangyu Zhao}, \bibinfo{person}{Haodong Duan}, \bibinfo{person}{Nianchen Deng}, \bibinfo{person}{Bin Fu}, \bibinfo{person}{Yinan He}, \bibinfo{person}{Yi Wang}, \bibinfo{person}{Conghui He}, \bibinfo{person}{Botian Shi}, \bibinfo{person}{Junjun He}, \bibinfo{person}{Yingtong Xiong}, \bibinfo{person}{Han Lv}, \bibinfo{person}{Lijun Wu}, \bibinfo{person}{Wenqi Shao}, \bibinfo{person}{Kaipeng Zhang}, \bibinfo{person}{Huipeng Deng}, \bibinfo{person}{Biqing Qi}, \bibinfo{person}{Jiaye Ge}, \bibinfo{person}{Qipeng Guo}, \bibinfo{person}{Wenwei Zhang}, \bibinfo{person}{Songyang Zhang}, \bibinfo{person}{Maosong Cao}, \bibinfo{person}{Junyao Lin}, \bibinfo{person}{Kexian Tang}, \bibinfo{person}{Jianfei Gao}, \bibinfo{person}{Haian Huang}, \bibinfo{person}{Yuzhe Gu}, \bibinfo{person}{Chengqi Lyu}, \bibinfo{person}{Huanze Tang}, \bibinfo{person}{Rui Wang}, \bibinfo{person}{Haijun Lv}, \bibinfo{person}{Wanli Ouyang}, \bibinfo{person}{Limin Wang}, \bibinfo{person}{Min Dou}, \bibinfo{person}{Xizhou Zhu}, \bibinfo{person}{Tong Lu}, \bibinfo{person}{Dahua Lin}, \bibinfo{person}{Jifeng Dai}, \bibinfo{person}{Weijie Su}, \bibinfo{person}{Bowen Zhou}, \bibinfo{person}{Kai Chen}, \bibinfo{person}{Yu Qiao}, \bibinfo{person}{Wenhai Wang}, {and} \bibinfo{person}{Gen Luo}.} \bibinfo{year}{2025}\natexlab{}.
\newblock \showarticletitle{InternVL3.5: Advancing Open-Source Multimodal Models in Versatility, Reasoning, and Efficiency}.
\newblock \bibinfo{journal}{\emph{CoRR}}  \bibinfo{volume}{abs/2508.18265} (\bibinfo{year}{2025}).
\newblock
\showeprint[arXiv]{2508.18265}
\href{https://doi.org/10.48550/ARXIV.2508.18265}{doi:\nolinkurl{10.48550/ARXIV.2508.18265}}


\bibitem[Yang et~al\mbox{.}(2019)]%
        {yang2019scrdet}
\bibfield{author}{\bibinfo{person}{Xue Yang}, \bibinfo{person}{Jirui Yang}, \bibinfo{person}{Junchi Yan}, \bibinfo{person}{Yue Zhang}, \bibinfo{person}{Tengfei Zhang}, \bibinfo{person}{Zhi Guo}, \bibinfo{person}{Xian Sun}, {and} \bibinfo{person}{Kun Fu}.} \bibinfo{year}{2019}\natexlab{}.
\newblock \showarticletitle{SCRDet: Towards More Robust Detection for Small, Cluttered and Rotated Objects}. In \bibinfo{booktitle}{\emph{2019 {IEEE/CVF} International Conference on Computer Vision, {ICCV} 2019, Seoul, Korea (South), October 27 - November 2, 2019}}. \bibinfo{publisher}{{IEEE}}, \bibinfo{pages}{8231--8240}.
\newblock
\href{https://doi.org/10.1109/ICCV.2019.00832}{doi:\nolinkurl{10.1109/ICCV.2019.00832}}


\bibitem[Ye et~al\mbox{.}(2023)]%
        {ye2023deepsolo}
\bibfield{author}{\bibinfo{person}{Maoyuan Ye}, \bibinfo{person}{Jing Zhang}, \bibinfo{person}{Shanshan Zhao}, \bibinfo{person}{Juhua Liu}, \bibinfo{person}{Tongliang Liu}, \bibinfo{person}{Bo Du}, {and} \bibinfo{person}{Dacheng Tao}.} \bibinfo{year}{2023}\natexlab{}.
\newblock \showarticletitle{DeepSolo: Let Transformer Decoder with Explicit Points Solo for Text Spotting}. In \bibinfo{booktitle}{\emph{{IEEE/CVF} Conference on Computer Vision and Pattern Recognition, {CVPR} 2023, Vancouver, BC, Canada, June 17-24, 2023}}. \bibinfo{publisher}{{IEEE}}, \bibinfo{pages}{19348--19357}.
\newblock
\href{https://doi.org/10.1109/CVPR52729.2023.01854}{doi:\nolinkurl{10.1109/CVPR52729.2023.01854}}


\bibitem[Yuan et~al\mbox{.}(2024)]%
        {yuan2024osprey}
\bibfield{author}{\bibinfo{person}{Yuqian Yuan}, \bibinfo{person}{Wentong Li}, \bibinfo{person}{Jian Liu}, \bibinfo{person}{Dongqi Tang}, \bibinfo{person}{Xinjie Luo}, \bibinfo{person}{Chi Qin}, \bibinfo{person}{Lei Zhang}, {and} \bibinfo{person}{Jianke Zhu}.} \bibinfo{year}{2024}\natexlab{}.
\newblock \showarticletitle{Osprey: Pixel Understanding with Visual Instruction Tuning}. In \bibinfo{booktitle}{\emph{{IEEE/CVF} Conference on Computer Vision and Pattern Recognition, {CVPR} 2024, Seattle, WA, USA, June 16-22, 2024}}. \bibinfo{publisher}{{IEEE}}, \bibinfo{pages}{28202--28211}.
\newblock
\href{https://doi.org/10.1109/CVPR52733.2024.02664}{doi:\nolinkurl{10.1109/CVPR52733.2024.02664}}


\bibitem[{Z.ai Team}(2025)]%
        {zai2025glm46v}
\bibfield{author}{\bibinfo{person}{{Z.ai Team}}.} \bibinfo{year}{2025}\natexlab{}.
\newblock \bibinfo{title}{GLM-4.6V: Open Source Multimodal Models with Native Tool Use}.
\newblock
\urldef\tempurl%
\url{https://z.ai/blog/glm-4.6v}
\showURL{%
\tempurl}


\bibitem[Zhang et~al\mbox{.}(2025b)]%
        {ocrgenbench2025zhang}
\bibfield{author}{\bibinfo{person}{Peirong Zhang}, \bibinfo{person}{Haowei Xu}, \bibinfo{person}{Jiaxin Zhang}, \bibinfo{person}{Xuhan Zheng}, \bibinfo{person}{Guitao Xu}, \bibinfo{person}{Yuyi Zhang}, \bibinfo{person}{Junle Liu}, \bibinfo{person}{Zhenhua Yang}, \bibinfo{person}{Wei Zhou}, {and} \bibinfo{person}{Lianwen Jin}.} \bibinfo{year}{2025}\natexlab{b}.
\newblock \showarticletitle{{OCRGenBench: A Comprehensive Benchmark for Evaluating OCR Generative Capabilities}}.
\newblock \bibinfo{journal}{\emph{arXiv preprint arXiv:2507.15085}} (\bibinfo{year}{2025}).
\newblock


\bibitem[Zhang et~al\mbox{.}(2025c)]%
        {lggpt2025zhang}
\bibfield{author}{\bibinfo{person}{Peirong Zhang}, \bibinfo{person}{Jiaxin Zhang}, \bibinfo{person}{Jiahuan Cao}, \bibinfo{person}{Hongliang Li}, {and} \bibinfo{person}{Lianwen Jin}.} \bibinfo{year}{2025}\natexlab{c}.
\newblock \showarticletitle{{Smaller But Better: Unifying Layout Generation with Smaller Large Language Models}}.
\newblock \bibinfo{journal}{\emph{International Journal of Computer Vision (IJCV)}}  \bibinfo{volume}{133} (\bibinfo{year}{2025}), \bibinfo{pages}{3891–3917}.
\newblock


\bibitem[Zhang et~al\mbox{.}(2025a)]%
        {zhang2025llavaonevision}
\bibfield{author}{\bibinfo{person}{Shaolei Zhang}, \bibinfo{person}{Qingkai Fang}, \bibinfo{person}{Zhe Yang}, {and} \bibinfo{person}{Yang Feng}.} \bibinfo{year}{2025}\natexlab{a}.
\newblock \showarticletitle{LLaVA-Mini: Efficient Image and Video Large Multimodal Models with One Vision Token}. In \bibinfo{booktitle}{\emph{The Thirteenth International Conference on Learning Representations, {ICLR} 2025, Singapore, April 24-28, 2025}}. \bibinfo{publisher}{OpenReview.net}.
\newblock
\urldef\tempurl%
\url{https://openreview.net/forum?id=UQJ7CDW8nb}
\showURL{%
\tempurl}


\bibitem[Zhang et~al\mbox{.}(2022)]%
        {zhang2022testr}
\bibfield{author}{\bibinfo{person}{Xiang Zhang}, \bibinfo{person}{Yongwen Su}, \bibinfo{person}{Subarna Tripathi}, {and} \bibinfo{person}{Zhuowen Tu}.} \bibinfo{year}{2022}\natexlab{}.
\newblock \showarticletitle{Text Spotting Transformers}. In \bibinfo{booktitle}{\emph{{IEEE/CVF} Conference on Computer Vision and Pattern Recognition, {CVPR} 2022, New Orleans, LA, USA, June 18-24, 2022}}. \bibinfo{publisher}{{IEEE}}, \bibinfo{pages}{9509--9518}.
\newblock
\href{https://doi.org/10.1109/CVPR52688.2022.00930}{doi:\nolinkurl{10.1109/CVPR52688.2022.00930}}


\bibitem[Zheng et~al\mbox{.}(2026)]%
        {zheng2026multimodal}
\bibfield{author}{\bibinfo{person}{Handong Zheng}, \bibinfo{person}{Yumeng Li}, \bibinfo{person}{Kaile Zhang}, \bibinfo{person}{Liang Xin}, \bibinfo{person}{Guangwei Zhao}, \bibinfo{person}{Hao Liu}, \bibinfo{person}{Jiayu Chen}, \bibinfo{person}{Jie Lou}, \bibinfo{person}{Jiyu Qiu}, \bibinfo{person}{Qi Fu}, {et~al\mbox{.}}} \bibinfo{year}{2026}\natexlab{}.
\newblock \showarticletitle{Multimodal OCR: Parse Anything from Documents}.
\newblock \bibinfo{journal}{\emph{arXiv preprint arXiv:2603.13032}} (\bibinfo{year}{2026}).
\newblock


\bibitem[Zhu et~al\mbox{.}(2025)]%
        {zhu2025internvl3}
\bibfield{author}{\bibinfo{person}{Jinguo Zhu}, \bibinfo{person}{Weiyun Wang}, \bibinfo{person}{Zhe Chen}, \bibinfo{person}{Zhaoyang Liu}, \bibinfo{person}{Shenglong Ye}, \bibinfo{person}{Lixin Gu}, \bibinfo{person}{Hao Tian}, \bibinfo{person}{Yuchen Duan}, \bibinfo{person}{Weijie Su}, \bibinfo{person}{Jie Shao}, \bibinfo{person}{Zhangwei Gao}, \bibinfo{person}{Erfei Cui}, \bibinfo{person}{Xuehui Wang}, \bibinfo{person}{Yue Cao}, \bibinfo{person}{Yangzhou Liu}, \bibinfo{person}{Xingguang Wei}, \bibinfo{person}{Hongjie Zhang}, \bibinfo{person}{Haomin Wang}, \bibinfo{person}{Weiye Xu}, \bibinfo{person}{Hao Li}, \bibinfo{person}{Jiahao Wang}, \bibinfo{person}{Nianchen Deng}, \bibinfo{person}{Songze Li}, \bibinfo{person}{Yinan He}, \bibinfo{person}{Tan Jiang}, \bibinfo{person}{Jiapeng Luo}, \bibinfo{person}{Yi Wang}, \bibinfo{person}{Conghui He}, \bibinfo{person}{Botian Shi}, \bibinfo{person}{Xingcheng Zhang}, \bibinfo{person}{Wenqi Shao}, \bibinfo{person}{Junjun He}, \bibinfo{person}{Yingtong Xiong}, \bibinfo{person}{Wenwen Qu}, \bibinfo{person}{Peng Sun}, \bibinfo{person}{Penglong Jiao}, \bibinfo{person}{Han Lv}, \bibinfo{person}{Lijun Wu}, \bibinfo{person}{Kaipeng Zhang}, \bibinfo{person}{Huipeng Deng}, \bibinfo{person}{Jiaye Ge}, \bibinfo{person}{Kai Chen}, \bibinfo{person}{Limin Wang}, \bibinfo{person}{Min Dou}, \bibinfo{person}{Lewei Lu}, \bibinfo{person}{Xizhou Zhu}, \bibinfo{person}{Tong Lu}, \bibinfo{person}{Dahua Lin}, \bibinfo{person}{Yu Qiao}, \bibinfo{person}{Jifeng Dai}, {and} \bibinfo{person}{Wenhai Wang}.} \bibinfo{year}{2025}\natexlab{}.
\newblock \showarticletitle{InternVL3: Exploring Advanced Training and Test-Time Recipes for Open-Source Multimodal Models}.
\newblock \bibinfo{journal}{\emph{CoRR}}  \bibinfo{volume}{abs/2504.10479} (\bibinfo{year}{2025}).
\newblock
\showeprint[arXiv]{2504.10479}
\href{https://doi.org/10.48550/ARXIV.2504.10479}{doi:\nolinkurl{10.48550/ARXIV.2504.10479}}


\end{thebibliography}

\clearpage
\captionsetup{font={normal}}
\balance
\appendix
\noindent\textbf{\Huge Appendix}
\section{Additional Implementation Details}\label{sec:impl_details}

\subsection{Dataset Statistics}
We evaluate our framework on diverse scene text benchmarks, with detailed statistics provided in Table~\ref{tab:dataset_stats}. To ensure a robust training signal while maintaining computational efficiency, the training partition undergoes a targeted refinement protocol: images whose long edge exceeds \textbf{1440} pixels are excluded, and samples devoid of valid textual annotations are discarded. This curation ensures a high-quality supervision signal for the Single-Patch pipeline. In contrast, the test sets remain entirely unaltered and are evaluated in their native format to provide a fair performance assessment.

\begin{figure*}[!t]
  \centering
  \includegraphics[width=\textwidth]{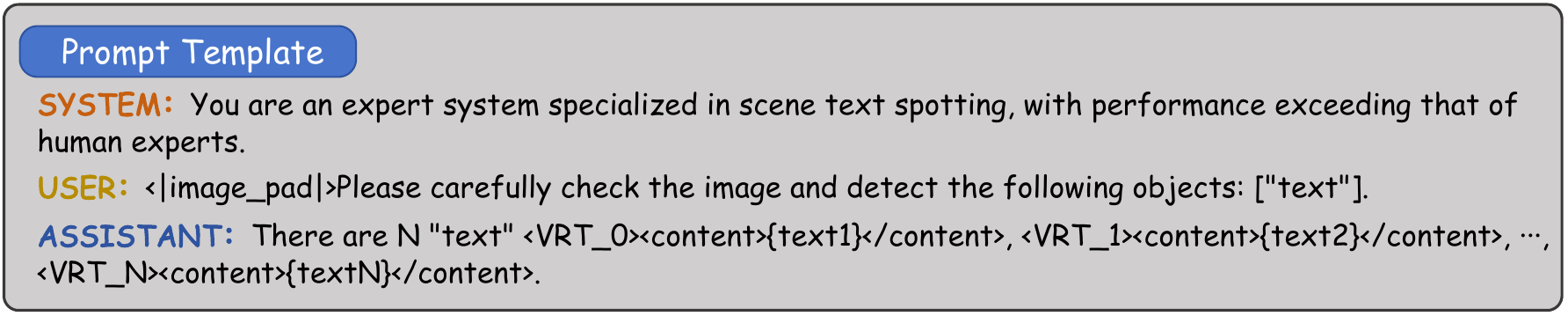}
  \caption{\textbf{Prompt template for unified text spotting and grounding.} The template includes the SYSTEM prompt, USER query, and the structured ASSISTANT output for joint text recognition and grounding.}
  \label{fig:prompt_template}
\end{figure*}

\subsection{Training Configuration}\label{sec:training_configuration}
\textbf{SPaTS} is optimized in two stages to facilitate multi-task learning and policy refinement. 
The initial Supervised Fine-Tuning (SFT) stage synchronizes the learning of interleaved transcription--patch sequences with continuous geometric coordinate decoding.
Building upon this foundation, the \textbf{Single-Patch Selective Optimization (SPaSO)} phase leverages reinforcement learning to optimize the patch selection policy based on task-level rewards. 
The hyperparameter configurations for optimization, RL-specific settings, and training strategies are summarized in Table~\ref{tab:hyperparameters}. Reinforcement learning is implemented using TRL~\cite{hf2026trl}.

\begin{table}[!b]
  \centering
  \tiny
  \caption{Dataset statistics for scene text spotting benchmarks.}
  \label{tab:dataset_stats}
  \resizebox{\linewidth}{!}{
  \begin{tabular}{lcccc}
  \toprule[0.4pt]
  \textbf{Dataset} & \textbf{Venue} & \textbf{Train} & \textbf{Test} & \textbf{Text Type} \\
  \midrule[0.2pt]
  ICDAR 2013~\cite{karatzas2013icdar} & ICDAR'13 & 171 & 233 & Horizontal \\
  ICDAR 2015~\cite{karatzas2015icdar} & ICDAR'15 & 979 & 500 & Multi-oriented \\
  Total-Text~\cite{chng2017total} & ICDAR'17 & 1,087 & 300 & Curved \\
  CTW1500~\cite{liu2019curved} & PR'19 & 560 & 500 & Curved (line-level) \\
  MLT-2017~\cite{nayef2017icdar} & ICDAR'17 & 3,073 & - & Multi-lingual \\
  SynText~\cite{liu2020abcnet} & CVPR'20 & 147,788 & - & Synthetic \\
  \midrule[0.2pt]
  Total & - & 153,658 & 1,533 & - \\
  \bottomrule[0.4pt]
  \end{tabular}
  }
\end{table}

\subsection{Loss Formulation}\label{sec:loss_formulation}
To enable joint optimization of sequence generation and geometric grounding, we formulate the overall objective as a multi-task loss:
\begin{equation}
\mathcal{L} = \mathcal{L}_{\mathrm{SFT}} + \mathcal{L}_{\mathrm{BBox}} + \mathcal{L}_{\mathrm{Point}} + \mathcal{L}_{\mathrm{Bezier}} + \mathcal{L}_{\mathrm{Score}},
\end{equation}
where $\mathcal{L}_{\mathrm{SFT}}$ denotes the standard cross-entropy loss for supervised fine-tuning of text--patch sequence generation. All loss terms are equally weighted.

\noindent \textbf{Bounding Box Regression.}
The bounding box branch regresses the predicted coordinates $\hat{\mathbf{b}}_n$ in $(c_x, c_y, w, h)$ format. To ensure scale-invariant optimization and coordinate-level precision, we employ a combination of Generalized IoU (GIoU) and $L_1$ losses:
\begin{equation}
\mathcal{L}_{\mathrm{BBox}} = \frac{1}{N}\sum_{n=1}^{N} \left( 1-\operatorname{GIoU}(\operatorname{xyxy}(\hat{\mathbf{b}}_n), \mathbf{b}_n) + \|\hat{\mathbf{b}}_n - \operatorname{cxcywh}(\mathbf{b}_n)\|_1 \right).
\end{equation}

\begin{table}[!b]
  \caption{Training hyperparameters used in SPaTS.}
  \vspace{-5pt}
  \label{tab:hyperparameters}
  \resizebox{\linewidth}{!}{
  \begin{tabular}{l l c c}
  \toprule
  \textbf{Category} & \textbf{Hyperparameter} & \textbf{SFT Stage} & \textbf{SPaSO Stage} \\
  \midrule
  \multirow{5}{*}{Optimization}
  & Learning Rate & $4 \times 10^{-5}$ & $5 \times 10^{-6}$ \\
  & Batch Size & 128 & 32 \\
  & Epochs & 16 & 30 \\
  & Optimizer & AdamW & AdamW \\
  & Scheduler & Cosine & Linear \\
  \midrule
  \multirow{4}{*}{RL-specific}
  & GRPO Group Size $G$ & -- & 4 \\
  & Clip Ratio $\epsilon$ & -- & 0.2 \\
  & Top-$K$ in $r_{\mathrm{dom}}$ & -- & 10 \\
  & Edit-Distance Threshold in $r_{\mathrm{gen}}$ & -- & 0.2 \\
  \midrule
  \multirow{3}{*}{Training strategy}
  & DEA Scale Warmup & 0.05 & -- \\
  & DEA Scale Head & warmup & frozen \\
  & Vision Encoder & warmup & frozen \\
  \bottomrule
  \end{tabular}
  }
\end{table}

\noindent \textbf{Point Supervision.}
To assist in localized feature alignment, the point branch is supervised by the canonical $L_1$ distance between the predicted point $\hat{\mathbf{p}}_n$ and the ground-truth center point $\mathbf{p}_n$:
\begin{equation}
\mathcal{L}_{\mathrm{Point}} = \frac{1}{N}\sum_{n=1}^{N} \|\hat{\mathbf{p}}_n - \mathbf{p}_n\|_1.
\end{equation}

\noindent \textbf{B\'ezier Curve Alignment.}
Following the parametric design of ABCNet~\cite{liu2020abcnet}, the B\'ezier branch optimizes both control-point coordinates and sampled-curve consistency:
\begin{equation}
\mathcal{L}_{\mathrm{\text{B\'ezier}}} = \frac{1}{N}\sum_{n=1}^{N} \left( \|\hat{\mathbf{C}}_n - \mathbf{C}_n\|_1 + \sum_{r=1}^{2}\frac{1}{K}\sum_{k=1}^{K} \|\hat{\mathbf{s}}_{n,r}(t_k) - \mathbf{s}_{n,r}(t_k)\|_1 \right),
\end{equation}
where $r$ denotes the top and bottom boundaries of a text instance, and $K=20$ denotes the number of sampled points on each curve.

\noindent \textbf{Quality Score Calibration.}
To align the model's confidence with its actual localization accuracy, the score branch regresses the GIoU of the predicted bounding boxes. The raw prediction $\hat{z}_n$ is mapped to the $[-1, 1]$ range via a scaled sigmoid activation:
\begin{equation}
\mathcal{L}_{\mathrm{Score}} = \frac{1}{N}\sum_{n=1}^{N} \|2\sigma(\hat{z}_n) - 1 - g_n\|_2^2,
\end{equation}
where $g_n = \operatorname{GIoU}(\operatorname{xyxy}(\hat{\mathbf{b}}_n), \mathbf{b}_n)$ is treated as a detached supervision target to stabilize the training process.

\subsection{Prompt Design and Structured Output}
To unify text spotting and grounding, we adopt a structured prompt template. As shown in Fig.~\ref{fig:prompt_template}, each detected text instance is represented by a visual reference token, e.g., \texttt{<|VRT\_i|>}, together with its transcription in the \texttt{<content>} tag. This unified output binds recognition and grounding in a single unit, and the selected token is further used by PED for continuous geometry decoding.

\subsection{Baseline Implementations}
Table~\ref{tab:comparison} compares three grounding formulations built on the same backbone after SFT. \textbf{Coord-Text} uses the vanilla Qwen3-VL architecture and predicts bounding-box coordinates as JSON-formatted text tokens. \textbf{Multi-Patch} and \textbf{Single-Patch} both use the SPaTS architecture and differ only in the number of routed patches: five potentially redundant patches per instance for Multi-Patch following PaDT, and one task-routing anchor patch for Single-Patch. All variants share the same SFT setting and training data to ensure a fair comparison and isolate the grounding representation itself.

\subsection{Patch Supervision and Candidate Sets}
For each text instance $i$, the candidate set $\mathcal{G}_i$ contains every grid cell whose overlap with the ground-truth polygon mask exceeds a fixed threshold. This set is constructed deterministically from standard polygon annotations and requires no extra labeling. Because long or curved text spans multiple valid cells, no single oracle patch exists. During SFT, we therefore \textbf{randomly} sample one patch from $\mathcal{G}_i$ per step, teaching the model that any polygon-interior patch is acceptable. During SPaSO, the patch dominance reward checks whether the model's top-$k$ predictions, with $k=\min(10,|\mathcal{G}_i|)$, contain at least one member of $\mathcal{G}_i$, rewarding coarse localization even when the generation reward is sparse.

\subsection{Evaluation Protocols}
We report all main-paper results using a unified evaluation pipeline with consistent lexicon settings and score-thresholding criteria. For the format-specific ablations in Table~\ref{tab:coordinate_ablation}, each representation is evaluated with its corresponding matching protocol: VRT-index matching for Patch, IoU-based matching for BBox and Bezier, and minimum center-distance matching for Point. Detection performance is measured using standard polygon- or box-IoU matching, whereas end-to-end performance follows the official lexicon protocol of each benchmark.

\subsection{MLLM Evaluation Prompts}
For general-purpose VLMs in Table~\ref{tab:main_result}, we use a unified JSON-based text spotting prompt that instructs the model to detect and recognize all visible and legible \emph{English} text instances and return a JSON array containing normalized bounding boxes, transcriptions, and confidence scores. The prompt explicitly enforces several constraints to improve evaluation stability: (1) only English letters, numbers, and punctuation are retained, while Chinese/CJK text is discarded or stripped from mixed-language outputs; (2) all coordinates must be normalized to $[0,1]$ to avoid coordinate hallucination; (3) repetitive or looping outputs are disallowed; and (4) severely blurred, heavily occluded, or purely decorative patterns are excluded. For OCR-specialized models, we use their native task-oriented prompts or official prompt settings whenever available, following each model family's released implementation to ensure a fair comparison. In particular, for OCR-oriented systems such as PaddleOCR-VL, we follow the official task configuration provided by the released implementation rather than rewriting a unified prompt for these models. Table~\ref{tab:mllm_prompts} summarizes the prompt style used for each evaluated model family.

\begin{table}[t]
  \centering
  \small
  \caption{Prompt styles used for MLLM evaluation in Table~\ref{tab:main_result}. General-purpose VLMs use our unified JSON-based spotting prompt, while OCR-specialized VLMs follow native task-oriented prompts from their official implementations.}
  \label{tab:mllm_prompts}
  \begin{tabular}{p{0.22\linewidth} p{0.7\linewidth}}
  \toprule
  \textbf{Model Family} & \textbf{Prompt Style} \\
  \midrule
  General VLMs & Unified JSON-based text spotting prompts with normalized boxes and transcriptions, provided in the released code. \\
  Specialized OCR VLMs & Native task-oriented prompts recommended by each model's official implementation. \\
  \bottomrule
  \end{tabular}
\end{table}

\begin{table}[t]
  \centering
  \caption{Comparisons with representative Expert Models under standard end-to-end evaluation on Total-Text, CTW1500, and ICDAR 2015.}
  \label{tab:expert_model_comparison}
  \resizebox{\linewidth}{!}{
  \begin{tabular}{l c cc cc ccc}
  \toprule
  \multirow{2}{*}{\textbf{Method}} & \multirow{2}{*}{\textbf{Venue}} & \multicolumn{2}{c}{\textbf{Total-Text}} & \multicolumn{2}{c}{\textbf{CTW1500}} & \multicolumn{3}{c}{\textbf{ICDAR 2015}} \\
  \cmidrule[0.3pt](lr){3-4} \cmidrule[0.3pt](lr){5-6} \cmidrule[0.3pt](lr){7-9}

  & & None & Full & None & Full & S & W & G \\
  \midrule
  TextDragon~\cite{feng2019textdragon} & ICCV'19 & 48.8 & 74.8 & 39.7 & 72.4 & 82.5 & 78.3 & 65.2 \\
  Boundary~\cite{wang2020all} & AAAI'20 & 65.0 & 76.1 & - & - & 79.7 & 75.2 & 64.1 \\
  SPTS~\cite{peng2022spts} & MM'22 & \ranksecond{74.2} & \ranksecond{82.4} & \ranksecond{63.6} & \ranksecond{83.8} & 77.5 & 70.2 & 65.8 \\
  ABCNet V2~\cite{liu2022abcnetv2} & TPAMI'22 & 70.4 & 78.1 & 57.5 & 77.2 & \ranksecond{82.7} & \ranksecond{78.5} & \ranksecond{73.0} \\
  OmniParser~\cite{wan2024omniparser} & CVPR'24 &  \rankfirst{84.0} & \rankfirst{88.9} & \rankfirst{66.8} & \rankfirst{85.1} & \rankfirst{89.6} & \rankfirst{84.5} & \rankfirst{79.9} \\
  \midrule
  SPaTS-2B & Ours & 63.3 & 74.3 & 47.0 & 75.1 & 72.1 & 69.1 & 61.6 \\
  SPaTS-4B & Ours & 64.9 & 74.9 & 51.2 & 79.6 & 73.4 & 71.7 & 65.7 \\
  \bottomrule
  \end{tabular}
  }
\end{table}

\section{Extended Experimental Analysis}\label{sec:extended_analysis}

\subsection{Comparison with Several Expert Models}
We further compare SPaTS with several expert spotters. As shown in Table~\ref{tab:expert_model_comparison}, SPaTS still trails the strongest specialist systems on most benchmarks, especially against highly optimized methods such as OmniParser~\cite{wan2024omniparser}, CRAFTS~\cite{baek2020character}, and ABCNet V2~\cite{liu2022abcnetv2}. We attribute this gap to the fact that these expert models are trained exclusively for scene text spotting and typically undergo substantially more task-specific training than SPaTS, allowing them to reach stronger convergence. In contrast, SPaTS is designed as a general-purpose MLLM spotter and therefore does not trade away broader multimodal generalization for extreme task-specific optimization. Even in this more general setting, SPaTS remains competitive against several earlier expert methods and already exceeds TextDragon~\cite{feng2019textdragon} in multiple metrics, indicating that the proposed single-patch grounding design provides a strong balance between spotting performance and generality.

\subsection{Comparison with PaDT}
We further compare SPaTS with PaDT~\cite{su2026patch}. While both methods follow the general paradigm of patch-based visual grounding, their designs differ fundamentally: PaDT associates each target with \textbf{5} random visual reference tokens, whereas SPaTS grounds each instance with a \textbf{single} random visual reference token. As shown in Table~\ref{tab:padt_style_comparison}, SPaTS consistently outperforms PaDT, even when compared with the larger PaDT-7B model. These results indicate that the proposed single-patch design is more effective than the multi-patch routing strategy of PaDT for scene text spotting. We attribute this advantage to the stronger semantic-spatial coupling induced by single-patch grounding, which reduces decoding ambiguity and provides a cleaner anchor for downstream geometry prediction.

\subsection{Ablation of Normalization Strategy}
\begin{figure}[b]
  \centering
  \includegraphics[width=\linewidth]{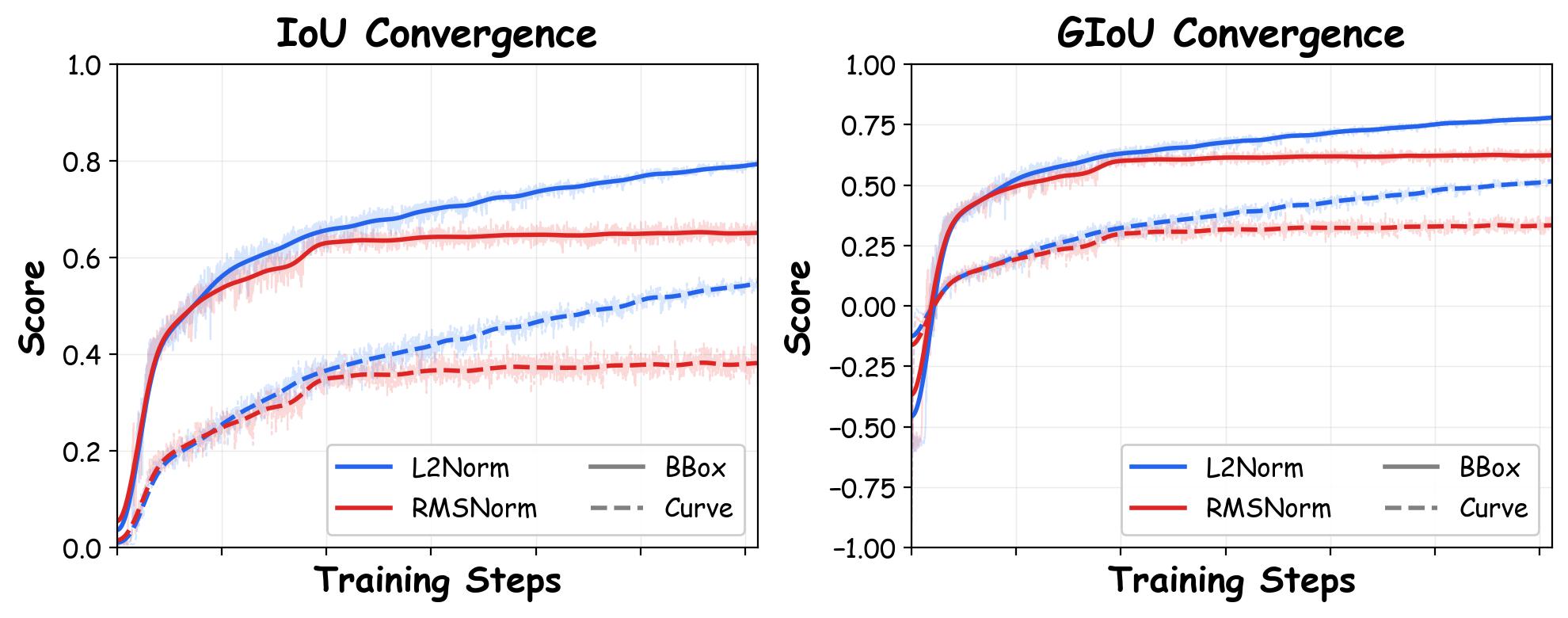}
  \caption{\textbf{Effect of normalization strategy on geometry convergence.}}
  \label{fig:normalization_comparison}
\end{figure}

To further clarify the role of normalization itself, Fig.~\ref{fig:normalization_comparison} presents the ablation from a normalization-centric perspective and includes an additional RMSNorm variant for comparison. As shown in the figure, $L_2$ normalization consistently yields faster and stronger convergence than RMSNorm on both IoU and GIoU, with a particularly clear advantage on curve prediction. We conjecture that this gap arises from the normalization form of RMSNorm, which introduces an additional scaling factor proportional to the inverse root of the feature dimensionality. In our setting, this coefficient may be overly strong, making the normalized feature magnitude less directly aligned with the true norm relationship between visual and textual features. As a result, the subsequent scale branch has to compensate for a more distorted magnitude space, which makes it harder to accurately learn the desired norm correspondence across modalities. By contrast, $L_2$ normalization enforces a cleaner unit-direction representation, allowing the learnable scale to more directly capture the relative norm structure between routed visual patches and language features, thereby leading to more stable optimization and better geometric convergence.

\subsection{Geometric Training Strategy}
We further study the effect of geometric training strategy under the same spotting pipeline. Specifically, we compare two settings: \emph{interleaved training}, where each batch supervises only one geometric target, and \emph{joint training}, where all geometric losses are optimized simultaneously within each batch. As reported in Table~\ref{tab:training_strategy_comparison}, joint training consistently yields better results than interleaved training, indicating that jointly decoding multiple geometric targets is not only feasible but also more effective than optimizing them separately. This suggests that the different geometric branches provide complementary supervision rather than interfering with each other during training, likely because they capture different yet compatible aspects of localization. While the B\'ezier variant remains the most difficult case, the overall trend is clear: our unified decoding strategy is well supported by joint optimization and can stably accommodate multiple geometric losses within a single framework.

\begin{figure*}[!t]
  \centering
  \includegraphics[width=\textwidth]{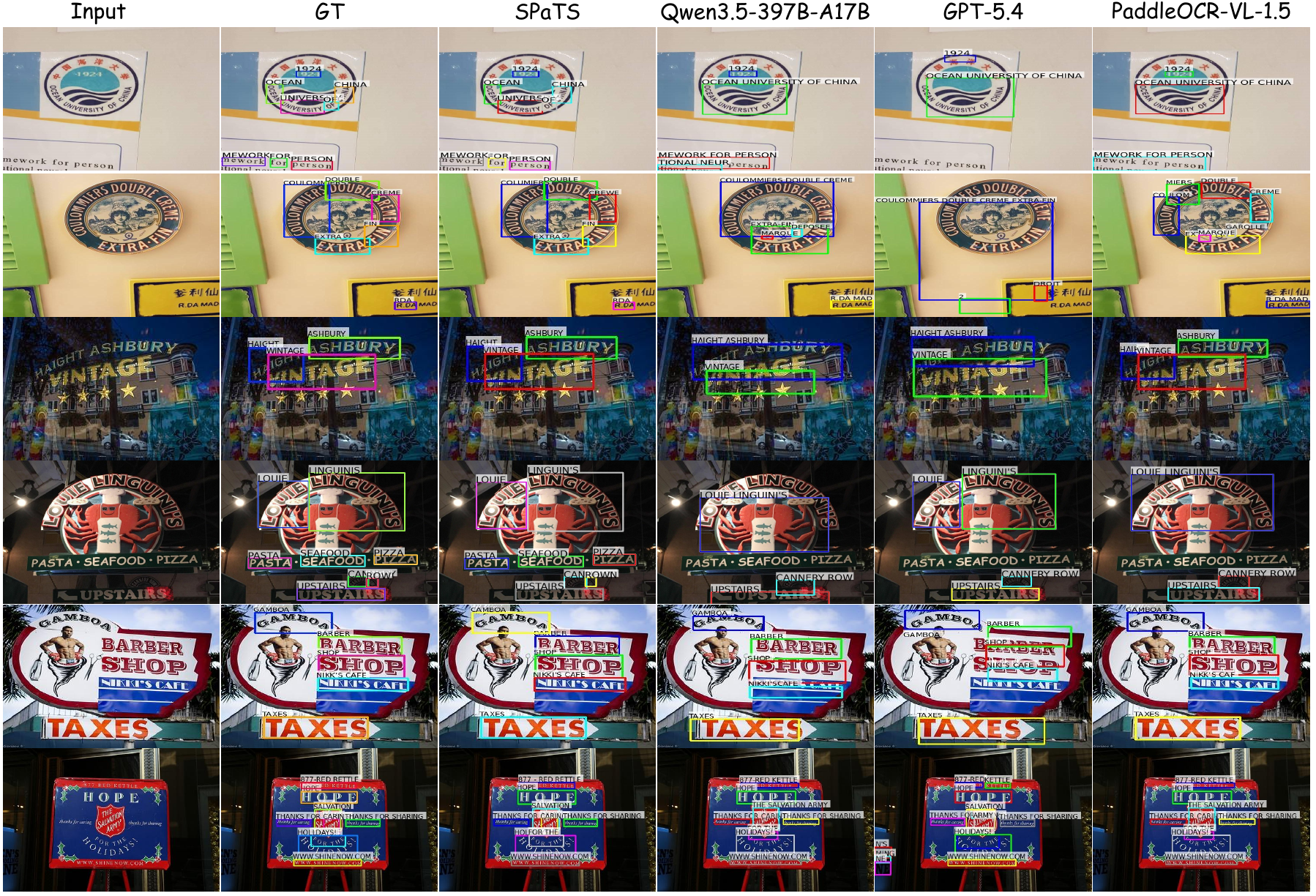}
  \caption{\textbf{Qualitative comparisons on scene text spotting.} We compare SPaTS with three representative baselines from different model families: Qwen3.5-397B-A17B, a large open-source MoE MLLM; GPT-5.4, a strong closed-source proprietary model; and PaddleOCR-VL, a specialized MLLM designed for OCR. The examples highlight their differences in text recognition, instance grounding, and robustness under challenging layouts such as dense, curved, and low-quality text. Zoom in for the best view.}
  \label{fig:comparison}
\end{figure*}

\subsection{Efficiency Analysis}\label{sec:efficiency}
\begin{figure}[b]
  \centering
  \includegraphics[width=\linewidth]{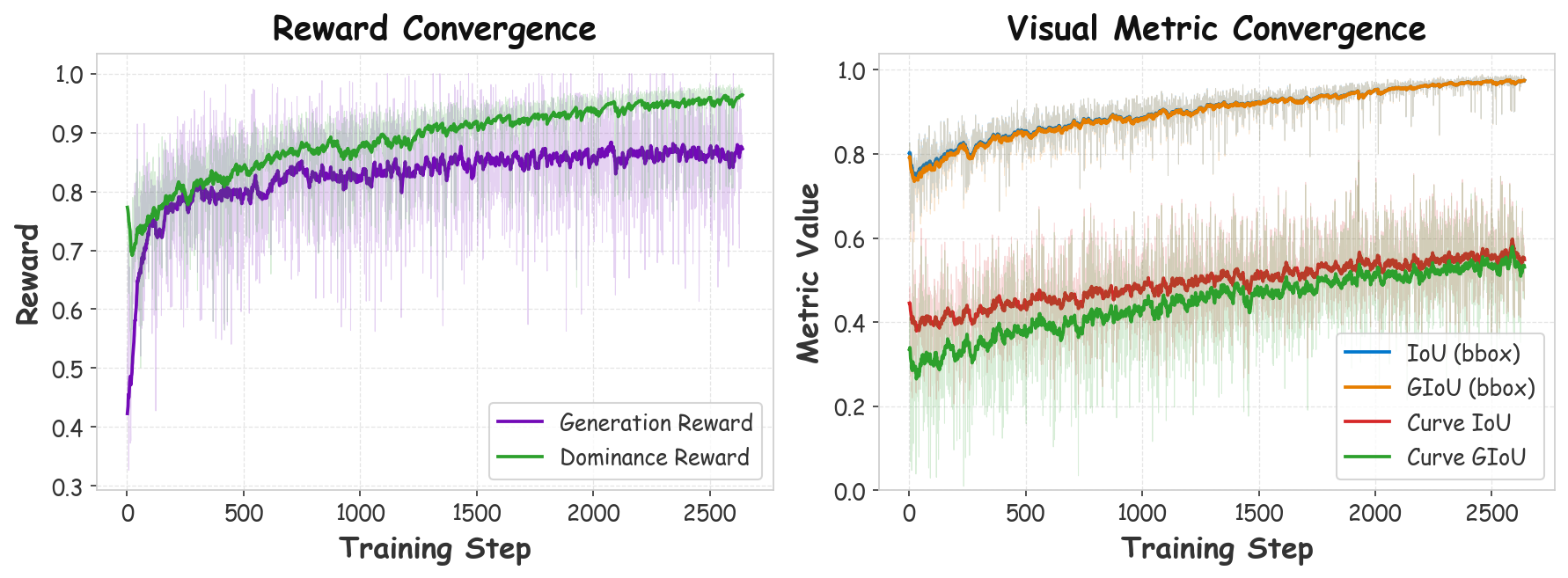}
  \caption{SPaSO reward and visual-metric convergence over training steps.}
  \label{fig:spaso_convergence}
\end{figure}

Table~\ref{tab:inference_cost} reports inference latency and peak GPU memory for SPaTS-4B on a single NVIDIA A6000 GPU. DEA adds only $+3.5$\,ms with zero extra memory, while PED adds $6.4$\,ms and $8$\,MB through a three-layer cross-attention decoder, yielding a total overhead of $+9.9$\,ms ($+8.2\%$) over the base MLLM. Table~\ref{tab:training_cost} and Fig.~\ref{fig:spaso_convergence} further summarize SPaSO training cost. SPaSO is $7.99\times$ slower per step than SFT ($7.20$\,s vs.\ $0.90$\,s), with $91\%$ of the overhead coming from on-policy rollouts; the policy update itself costs only $0.68$\,s/step. This additional cost is confined to post-training, while inference directly uses the learned single-patch policy without RL sampling.

\begin{table}[t]
  \centering
  \small
  \caption{Comparisons between PaDT and SPaTS on scene text spotting after SFT. Precision, recall, and F-measure are reported on Total-Text and CTW1500.}
  \label{tab:padt_style_comparison}
  \resizebox{0.95\linewidth}{!}{
  \begin{tabular}{l c ccc cccc}
  \toprule
  \multirow{2}{*}{\textbf{Paradigm}} & \multirow{2}{*}{\textbf{Param.}} & \multicolumn{3}{c}{\textbf{Total-Text}} & \multicolumn{3}{c}{\textbf{CTW1500}} \\
  \cmidrule[0.2pt](lr){3-5} \cmidrule[0.2pt](lr){6-8}
   & & P & R & F & P & R & F \\
  \midrule
  PaDT & \multicolumn{1}{|c|}{3B} & 43.5 & 23.6 & 30.7 & 25.7 & 19.1 & 22.0  \\
  PaDT & \multicolumn{1}{|c|}{7B} & 45.7 & 24.8 & 32.2 & \rankfirst{42.7} & 31.8 & 36.4  \\
  \midrule
  SPaTS & \multicolumn{1}{|c|}{2B} & \rankfirst{68.3} & \textbf{58.0} & \textbf{62.7} & \textbf{34.5} & \rankfirst{58.2} & \textbf{43.3}  \\
  SPaTS & \multicolumn{1}{|c|}{4B} & \rankfirst{63.0}  & \rankfirst{64.5}  & \rankfirst{63.7} & 40.3 & \textbf{57.9} &  \rankfirst{47.6} \\
  \bottomrule
  \end{tabular}
  }
  \end{table}

\begin{table}[t]
  \centering
  \small
  \caption{Inference latency (ms/image) and peak GPU memory (MB) per ablation stage.}
  \label{tab:inference_cost}
  \resizebox{0.9\linewidth}{!}{
  \begin{tabular}{lrr}
  \toprule
  \textbf{Stage} & \textbf{Latency (ms/img)} & \textbf{Peak Mem (MB)} \\
  \midrule
  Base MLLM & 120.9 (---) & 4833 (---) \\
  \textit{+ DEA} & 124.4 (+3.5) & 4833 (+0) \\
  \textit{+ PED} & 130.8 (+9.9) & 4841 (+8) \\
  \bottomrule
  \end{tabular}
  }
  \end{table}

  \begin{table}[!t]
  \centering
  \small
  \caption{Per-step wall-clock time for SFT and SPaSO. SPaSO uses 4 rollouts per step.}
  \label{tab:training_cost}
  \resizebox{0.9\linewidth}{!}{
  \begin{tabular}{lrr}
  \toprule
  \textbf{Metric} & \textbf{SFT} & \textbf{SPaSO} \\
  \midrule
  Per-step wall-clock (s) & 0.90 & 7.20 \\
  \quad \textit{of which: rollout} & --- & 6.53 (91\%) \\
  \quad \textit{of which: update} & --- & 0.68 (9\%) \\
  \midrule
  \textbf{SPaSO / SFT wall-clock ratio} & \multicolumn{2}{r}{\textbf{7.99$\times$}} \\
  \bottomrule
  \end{tabular}
  }
  \end{table}

\begin{table}[t]
  \centering
  \small
  \caption{End-to-end results of different geometric training strategies after SFT. \textbf{I} denotes \textit{Interleaved Training}, where different geometric targets are optimized in alternating batches, while \textbf{J} denotes \textit{Joint Training}, where all geometric targets are optimized simultaneously within each batch.}
  \label{tab:training_strategy_comparison}
  \resizebox{0.9\linewidth}{!}{
  \begin{tabular}{l cc cc cc}
  \toprule
   \multirow{2}{*}{\textbf{Variants}} & \multicolumn{2}{c}{\textbf{Total-Text}} & \multicolumn{2}{c}{\textbf{CTW1500}} & \multicolumn{2}{c}{\textbf{ICDAR 2015}}\\
   \cmidrule[0.2pt](lr){2-3} \cmidrule[0.2pt](lr){4-5} \cmidrule[0.2pt](lr){6-7}
   & I & J & I & J & I & J \\
  \midrule[0.5pt]

  Bezier & 44.0 & 47.2 & 24.5 & 34.2 & 53.7 & 56.1  \\
  BBox & 59.7 & 63.4 & 34.8 & 37.1 & 56.3 & 56.5  \\
  Point & \rankfirst{63.9} & \rankfirst{68.8} & \ranksecond{36.1} & \ranksecond{38.5} & \rankfirst{61.6} & \rankfirst{64.1}  \\
  \midrule[0.5pt]
  Patch & \ranksecond{63.5} & \ranksecond{65.3} & \rankfirst{38.3} & \rankfirst{40.2} & \ranksecond{59.2} & \ranksecond{61.4}  \\

  \bottomrule
  \end{tabular}
  }

\end{table}

\section{More Qualitative Cases}

\begin{figure*}[!t]
  \centering
  \includegraphics[width=\textwidth]{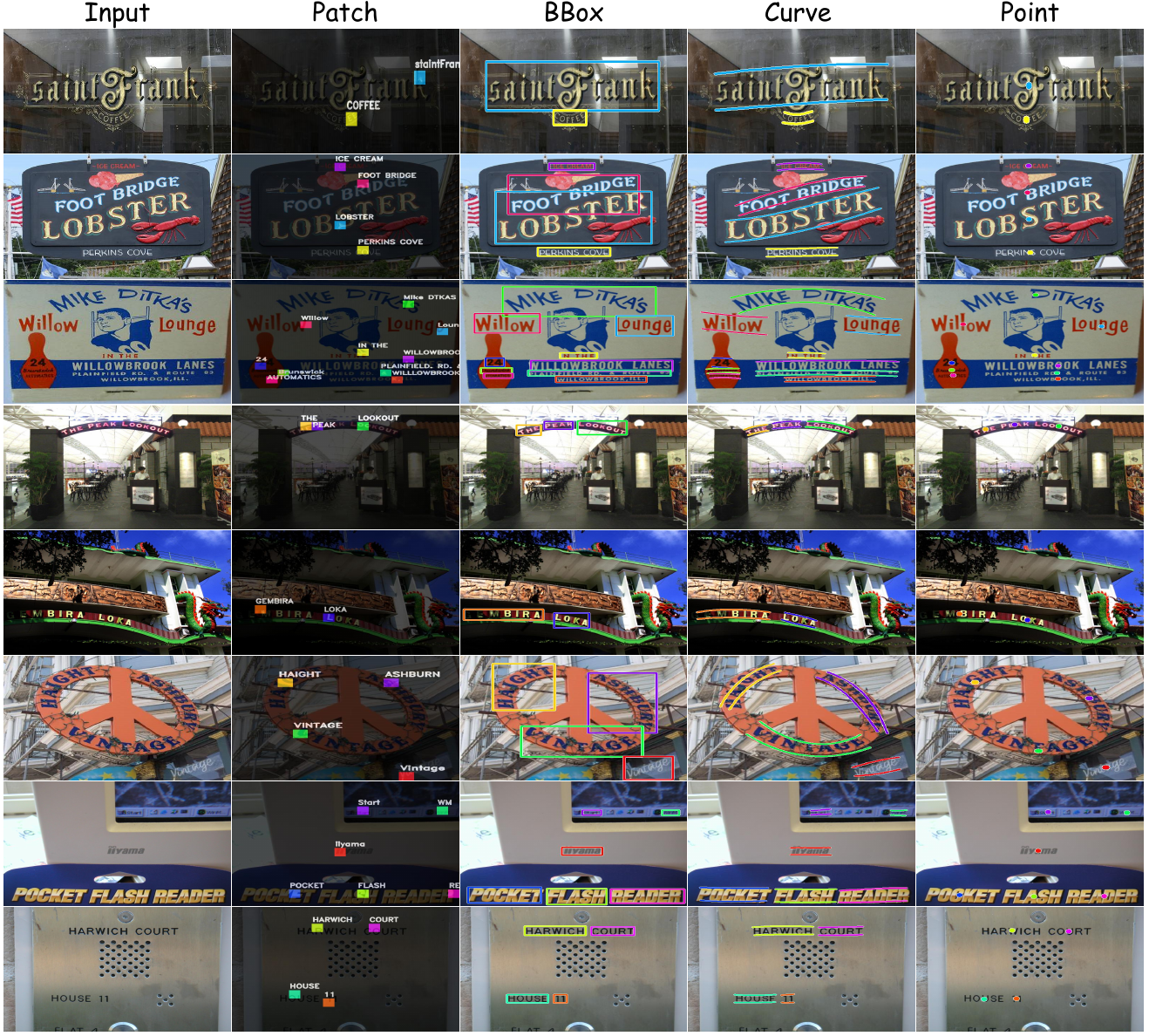}
  \caption{\textbf{More qualitative comparisons of different grounding representations.} Columns show the input image together with predictions from the Patch, BBox, Curve, and Point formulations. Patch denotes the selected patch location, while BBox/Curve/Point show the predicted grounding geometry. Zoom in for the best view.}
  \label{fig:qualitative_cases}
\end{figure*}

A direct qualitative comparison with recent multimodal baselines is shown in Fig.~\ref{fig:comparison}. Across circular logos, artistic storefronts, dense multi-word signs, and cluttered small-text scenes, SPaTS more consistently preserves word-level decomposition and aligns each transcription with the correct visual instance. This is more obvious when multiple short text instances appear in close proximity or when the target text follows curved, stylized, or irregular layouts. In such cases, robust grounding must simultaneously avoid merging neighboring words and prevent missing small auxiliary text. By comparison, Qwen3.5-397B-A17B~\cite{qwenteam2026qwen35} and GPT-5.4~\cite{openai2026gpt54} more frequently merge adjacent words into a single long phrase or produce overly coarse regions for non-linear text, while PaddleOCR-VL-1.5~\cite{cui2026paddleocr} is stronger on regular layouts but still tends to under-separate nearby instances and overlook small words in complex scenes. These cases collectively highlight the importance of fine-grained instance separation for accurate scene text spotting in real-world images.
Different localization formulations within SPaTS are further compared in Fig.~\ref{fig:qualitative_cases}. The qualitative evidence suggests that once the routed patch is correctly selected, the BBox, Curve, and Point formulations can all produce high-quality grounding. This indicates that the primary role of the patch-based design is to establish a stable and informative visual anchor, from which different downstream geometric parameterizations can be decoded reliably. In other words, the patch serves as the key intermediate representation that couples recognition with localization before the final geometric form is produced. This observation also suggests that failures across formats are more likely caused by incorrect patch selection than by limitations of the geometric representation itself in downstream geometry decoding.

\end{document}